\documentclass[letterpaper]{article} 
\usepackage{aaai25}  
\usepackage{times}  
\usepackage{helvet}  
\usepackage{courier}  
\usepackage[hyphens]{url}  
\usepackage{graphicx} 
\usepackage{natbib}  
\usepackage{caption} 
\usepackage{algorithm}
\usepackage{algorithmic}

\usepackage{newfloat}
\usepackage{listings}

\usepackage{amsfonts}
\usepackage[inkscapelatex=false]{svg}
\usepackage{amsmath}

\usepackage{adjustbox}
\usepackage{booktabs}
\usepackage{subcaption}

\DeclareCaptionStyle{ruled}{labelfont=normalfont,labelsep=colon,strut=off} 
\floatstyle{ruled}
\newfloat{listing}{tb}{lst}{}
\floatname{listing}{Listing}
\title{Teaching AI the Anatomy Behind the Scan: \\ Addressing Anatomical Flaws in Medical Image Segmentation with \\
Learnable Prior}
\author{
    Jeon Young Seok \textsuperscript{\rm 1}\equalcontrib,
    Hongfei Yang \textsuperscript{\rm 1}\equalcontrib,
    Huazhu Fu \textsuperscript{\rm 2},
    Mengling Feng \textsuperscript{\rm 1}\thanks{corresponding}
}
\affiliations{
    \textsuperscript{\rm 1} National University of Singapore, Singapore \\
    \textsuperscript{\rm 2} Agency for Science, Technology and Research (A*STAR), Singapore
}

\begin{document}

\maketitle

\begin{abstract}

Imposing key anatomical features, such as the number of organs, their shapes and relative positions, is crucial for building a robust multi-organ segmentation model. Current attempts to incorporate anatomical features include broadening the effective receptive field (ERF) size with data-intensive modules, or introducing anatomical constraints that scales poorly to multi-organ segmentation.
%
We introduce a novel architecture called the Anatomy-Informed Cascaded Segmentation Network (AIC-Net). AIC-Net incorporates a learnable input termed ``Anatomical Prior'', which can be adapted to patient-specific anatomy using a differentiable spatial deformation. The deformed prior later guides decoder layers towards more anatomy-informed predictions. We repeat this process at a local patch level to enhance the representation of intricate objects, resulting in a cascaded network structure.
%
AIC-Net is a general method that enhances any existing segmentation models to be more anatomy-aware. We have validated the performance of AIC-Net, with various backbones, on two multi-organ segmentation tasks: abdominal organs and vertebrae. For each respective task, our benchmarks demonstrate improved dice score and Hausdorff distance.
%
\end{abstract}

%

\section{Introduction}

It is becoming increasingly common to encounter AI models with reported performance on par with, or even surpassing, radiologists in various medical segmentation tasks \cite{hirsch2021radiologist}. 
However, it is highly unlikely that these AI models will replace radiologists anytime soon \cite{waymel2019impact}.
Although the models report good statistical results, examining each case frequently uncovers anatomically flawed predictions that radiologists would never make.
In bone segmentation, AI can confuse nearby vertebrae as they appear similar locally, leading to mixed predictions. 
In abdominal organ segmentation tasks, the AI could incorrectly detect the esophagus, a muscular tube that carries food from the mouth to the stomach, resulting in fragmented predictions. 
These examples demonstrate that current segmentation models do not reason in the same way that radiologists do; who has a comprehensive understanding of human anatomy, which enables them to make a more anatomy-informed judgements, whereas AI models seem to struggle to grasp such concepts.

\begin{figure}[t!]
     \centering
     \begin{subfigure}[b]{0.45\columnwidth}
         \centering
         \includegraphics[width=\textwidth]{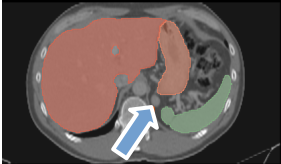}
         \caption{}
         \label{fig:311_2d}
     \end{subfigure}
     \begin{subfigure}[b]{0.45\columnwidth}
         \centering
         \includegraphics[width=\textwidth]{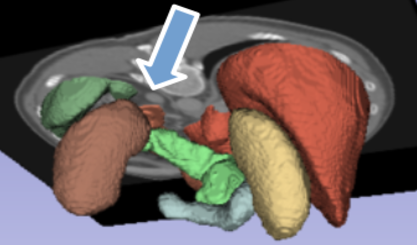}
         \caption{}
         \label{fig:311_2d_3d_fuse}
     \end{subfigure}
     \begin{subfigure}[b]{0.45\columnwidth}
         \centering
         \includegraphics[width=\textwidth]{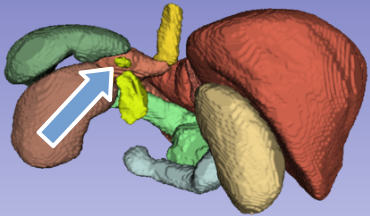}
         \caption{}
         \label{fig:unet_311}
     \end{subfigure}
     \begin{subfigure}[b]{0.45\columnwidth}
         \centering
         \includegraphics[width=\textwidth]{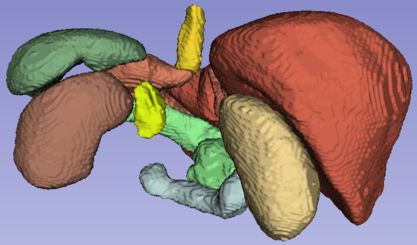}
         \caption{}
         \label{fig:aic_net_311}
     \end{subfigure}
     \hfill
    \caption{Shall we label the gray spot indicated by the blue arrow adrenal gland? (a) scan slice, (b) ground truth (with adrenal gland label removed) 3D segmentation around the slice, (c) all baseline segmentation wrongly segmented the spot as gland, and (d) AIC-Net gives correct segmentation.}
    \label{fig:local_feature_not_enough}
\end{figure}

So, what causes existing segmentation models to find it difficult to recognize anatomical features, which humans can grasp from just a handful of examples, despite being trained on hundreds of thousands of instances?
AI-driven segmentation models are trained to detect organs solely from CT/MRI scans~\cite{cciccek20163d,chen2018encoder,hatamizadeh2021swin}. Ideally, a robust model should extract both local and global features, using global features to distinguish similar-looking local features. However, we often observe that when relying solely on the scan as input, these models tend to overlook learning global patterns.
%
%
For example, in the scan slice shown in Figure~\ref{fig:311_2d}, base on local patterns it is difficult to tell if the gray spot, as indicated by the blue arrow, should be segmented as left adrenal gland or not. It is positioned directly above the right kidney, where the gland typically appears, and has similar intensities to the average adrenal gland. All baseline models we tested wrongly segmented it as part of the left adrenal gland, as shown in Figure~\ref{fig:unet_311}. However, this results in a separated component of the gland, clearly violating the anatomy. Our proposed method can give correct segmentation.

Several methods to enhance AI's understanding of anatomical features in medical segmentation can be grouped into two main approaches: 1) Expanding the model's search scope by utilizing broader effective receptive fields (ERF)~\cite{luo2016understanding}, and 2) Constraining predictions through regularized cost functions or incorporating prior knowledge.
%
In widening ERF, numerous studies have investigated replacing convolutional blocks with other computational blocks with a broader ERF, such as Transformers~\cite{vaswani2017attention, hatamizadeh2021swin} and State Space Models~\cite{gu2023mamba, wang2024mamba}. While models with larger ERFs are generally more adept at identifying global features, they often require more training data to achieve better generalization~\cite{dosovitskiy2020image}, which can be a significant bottleneck in the medical domain.
%
In adding constraints, several studies have used topological losses with persistent homology to impose topological constraints on predictions~\cite{santhirasekaram2023topology,hu2019topology}. While effective for single-object predictions, this method struggles with multi-organ segmentation, where organ shapes and relative locations are critical. Some works reformulate segmentation as a deformation problem, learning to warp a fixed template represented as a mesh~\cite{bongratz2023abdominal,kong2021deep} or pixels~\cite{wang2012multi,lee2019tetris}. While this yields smoother, noise-free predictions, it often struggles with small intricate structures and its prediction accuracy depends heavily on the template quality.

In this paper, we introduce a novel approach called Anatomically Informed Cascaded Segmentation Net (AIC-Net), which can be integrated with any standard segmentation network to ensure anatomically accurate predictions, without relying on data-intensive self-attention modules or template-matching approach that struggles in representing complex structures.
Instead, AIC-Net introduces a learnable parameter called \textit{Anatomical Prior}, which can be spatially deformed to align with the anatomy of a patient and serves as a soft constraint during prediction. Specifically, given a 3D scan, a portion of the encoder learns to predict the control parameters of affine and thin plate spline (TPS) spatial deformations~\cite{bookstein1989principal}. The deformation functions adjust the learnable prior to match the patient's anatomy. This deformed prior is then integrated during the decoding phase to guide the decoder towards more anatomically accurate predictions. To further enhance deformation accuracy for intricate structures, the process is repeated using cropped local patches, resulting in a global-local cascaded structure.

AIC-Net is a general method that enhances any existing segmentation model to be more anatomy-aware. We have validated the performance of AIC-Net on two segmentation tasks: abdominal organ and vertebrae from TotalSegmentator dataset~\cite{wasserthal2023totalsegmentator}. Our benchmarks consistently demonstrate improved performance with the addition of a learnable prior.

The contributions of this paper are summarized as follows:

\begin{itemize}
    \item We propose boosting the robustness of multi-organ segmentation models by introducing a learnable free parameter termed ``Anatomical Prior'' which learns a generic human anatomy. The prior serves as a soft constraint during decoding process. 
    
    \item The learned Anatomical Prior is tailored to match each patient's unique anatomy using deformation methods such as Thin-Plate Spline (TPS) and affine, enabling complex deformations with minimal control parameters. We further refine the details of the learned Anatomical Prior for intricate objects by repeating the process at a local patch level, resulting in a cascaded structure.

    \item We propose a novel centroid loss that encourages the alignment of centroids between the deformed Anatomical Prior and the ground truth, which is crucial for attaining a realistic prior.
\end{itemize}

\section{Prior Works}

Existing methods for enhancing anatomical feature learning focus on broadening ERF, reformulating segmentation to mesh deformation, or imposing topological constraints with regularizers, each with its own drawbacks.
\subsection{Broadening ERF}
self-attention networks~\cite{vaswani2017attention} can attain larger ERF than CNNs. Therefore, these models are more suitable for learning distant dependencies within the data, making them good candidates for learning anatomical feature~\cite{chen2021transunet,hatamizadeh2021swin,petit2021u}. However, in practice, these models may struggle to effectively learn anatomical priors due to the limited data available to supervise the learning of long-range dependencies.  
Numerous results show worse performance on transformer-based models when trained with limited data.~\cite{isensee2024nnu,luo2021word,roy2023mednext}.

\subsection{Mesh Deformation}
Mesh-deformation~\cite{kong2021deep,bongratz2023abdominal,dalca2019unsupervised,van2008encoding} computes a differentiable deformation of grid meshes to deform a predefined prior to fit specific scans.
The warped prior then can be used to produce segmentation.
This approach naturally offers smoother contour predictions compared to conventional pixel prediction, but it may encounter difficulties representing intricate structures. Moreover, it is challenging to estimate accurate and robust deformations, and failed estimations can cause flipping or self-intersection of predicted objects \cite{gao2020beyond}. One potential solution is integrating mesh-based segmentation with pixel-based methods. However, this approach poses challenges due to the differing nature of object representation between the two methods.

\subsection{Topology regularization}
The shapes of organs serve as crucial anatomical characteristics, which can be described by their topological features, such as the number of objects and cavities within 3D volumes. Persistent homology offers an efficient method to summarize these topological features across multiple resolutions  \cite{dey2022computational}.
Several studies have employed topological constraints to regularize network predictions~\cite{zhang2022progressive,hu2019topology,byrne2022persistent}.
Nevertheless, integrating topological features directly into a deep learning framework poses significant challenges due to their discrete nature, which complicates the gradient flow in neural networks. 
This complexity is further exacerbated in multi-organ segmentation, where topological features become increasingly intricate, making regularization even more difficult \cite{byrne2022persistent}.


\begin{figure*}[t]
    \centering
    \includegraphics[width=1.75\columnwidth]{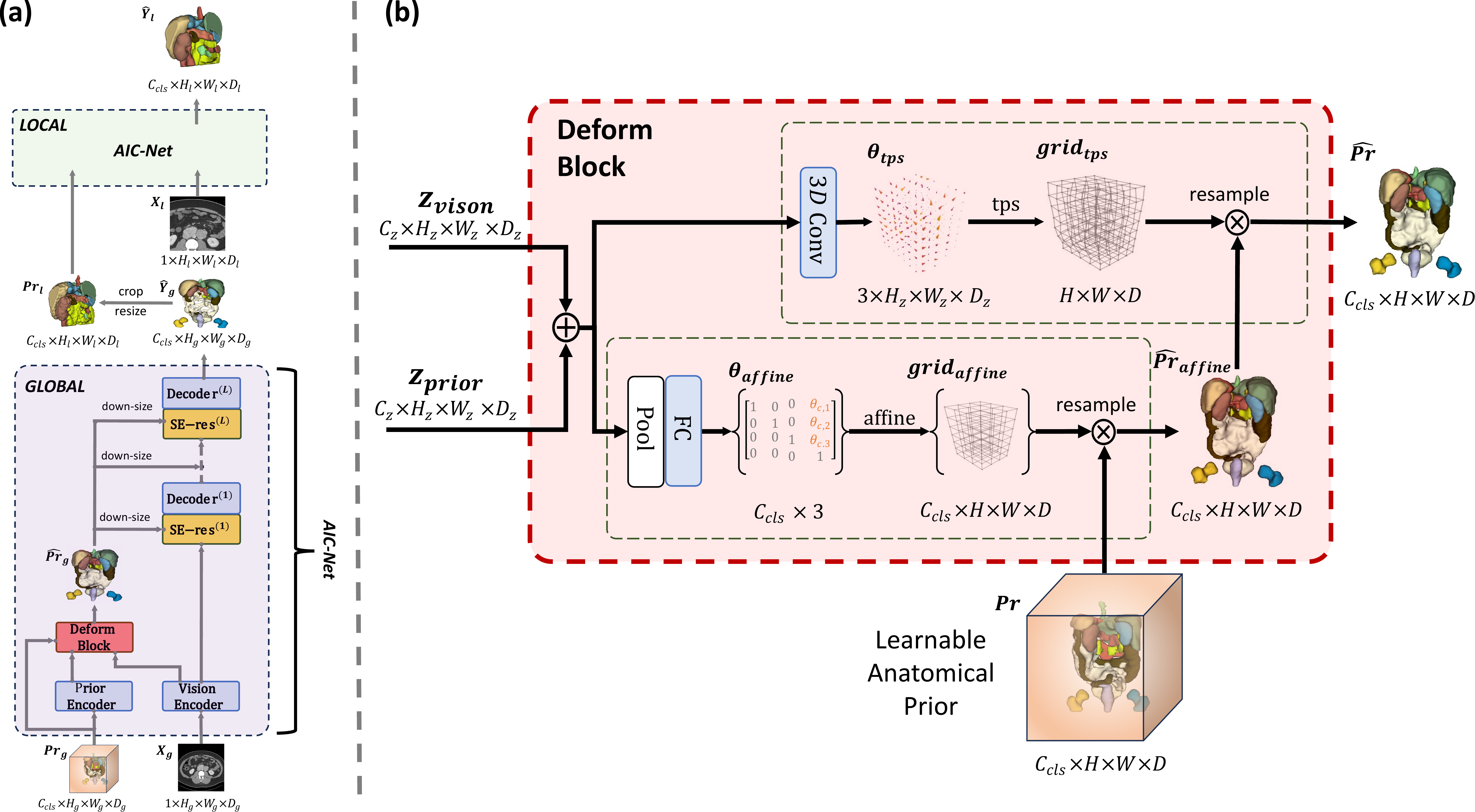}
    \caption{\textbf{(a)} Overview of AIC-Net. AIC-Net is a cascaded network combining global and local views for comprehensive multi-organ segmentation. Initial input $\mathbf{X}_{g}$ yields rough global prediction $\widehat{\mathbf{Y}}{g}$, enhanced by a learnable Anatomical Prior $\widehat{\mathbf{Pr}}_{g}$, a spatially deformed anatomy from learnable parameters $\mathbf{Pr}_{g}$ via $\text{Deform}_g$. This process repeats in the local segment of the model for further enhancements, taking local view $\mathbf{X}_{l}$ and local prior $\mathbf{Pr}_{l}$. \textbf{(b)} The Deform block receives embeddings from vision and prior encoders, concatenates them, and performs affine and TPS deformations on Anatomical Prior. Affine translates each organ. TPS warps the translated organ for more precise matching.}
    \label{fig:model}
\end{figure*}

\section{Method}
\subsection{Network Overview}
\sloppy AIC-Net (depicted in Figure~\ref{fig:model}) is a cascaded network that utilizes both a global view $\mathbf{X}_{g} \in \mathbb{R}^{1 \times H_g\times W_g \times D_g}$ and a local view $\mathbf{X}_{l} \in \mathbb{R}^{1 \times H_l\times W_l \times D_l}$ to produce a comprehensive local multi-organ prediction $\widehat{\mathbf{Y}}_{l} \in [0,1]^{C_\text{cls}\times H_l\times W_l \times D_l}$ as the final output, where $C_\text{cls}$ signifies the number of organs. 
%

At a global level, AIC-Net begins by taking $\mathbf{X}_g$, a down-sampled view of a raw scan, as input to generate a global prediction $\widehat{\mathbf{Y}}_g \in [0,1]^{C_\text{cls} \times H_g \times W_g \times D_g}$.
In producing $\widehat{\mathbf{Y}}_g$, alongside the standard encoder-decoder architecture, AIC-Net introduces a learnable parameter termed ``Anatomical Prior'' $\mathbf{Pr}_g \in \mathbb{R}^{C_\text{cls} \times H_g \times W_g \times D_g}$ as well as three types of computational blocks: $\text{PriorEncoder}_g$, $\text{Deform}_g$, and $\{\text{SE-res}_g^{(i)}\}$.

Given a prior $\mathbf{Pr}_g$ that is optimized to represent a generic anatomy, the deformation module $\text{Deform}_g$ deforms $\mathbf{Pr}_g$ into a deformed prior $\widehat{\mathbf{Pr}}_g$ that matches the anatomy of the given scan $\mathbf{X}_g$.
The extent of deformation is learned by the features from the vision encoder $\text{Encoder}_g$ and a lightweight prior encoder module $\text{PriorEncoder}_g$.
The deformed prior $\widehat{\mathbf{Pr}}_g$ is subsequently combined with the intermediate features from each of the decoder blocks $\{\text{Decoder}_g^{(l)}\}$ via the feature aggregation modules $\{\text{SE-res}_g^{(l)}\}$, guiding the decoder blocks to produce anatomy-informed predictions.

This process repeats in the local segment of the model, taking the local view $\mathbf{X}_l$ and the local Anatomical Prior $\mathbf{Pr}_l$, which is cropped and up-sized from $\widehat{\mathbf{Y}}_g$, as input. The local model serves to refine the global deformed prior, producing a deformed local prior $\widehat{\mathbf{Pr}}_l$.


\subsection{Deform block}

As illustrated in Figures~\ref{fig:model}, the $\text{Deform}$ block receives two embeddings, $\mathbf{z}_{\text{vision}} \in \mathbb{R}^{C_z \times H_z \times W_z \times D_z}$ from the vision encoder and $\mathbf{z}_{\text{prior}} \in \mathbb{R}^{C_z \times H_z \times W_z \times D_z}$ from the prior encoder, as inputs.
Within the Deform block, the two inputs are concatenated to create a single embedding. 
This unified embedding is then utilized to execute two types of spatial deformation: 1) affine and 2) TPS deformation~\cite{bookstein1989principal}.
Both deformation techniques are differentiable, enabling gradient-based optimization.
Prior $Pr$ first goes through the affine transform and thereafter the TPS.
The affine transform translates each organ in prior $Pr$ to align their centroids with the organs in the given scan.
In contrast, the goal of TPS is to warp the center-aligned organs in a non-linear fashion to match their shapes.

\subsubsection{Affine block}
As illustrated in Fig~\ref{fig:model}(b), the affine deform block initially performs a global pooling to the concatenated 3D embedding.
Subsequently, an FC layer maps the pooled embedding to the size of $C_{\text{cls}}\times 3$, corresponding to the affine transformation parameters for $C_{\text{cls}}$ organs along the $h$-, $w$-, and $z$-axes.
For a given target coordinate $\textbf{p} = (x, y, z)$, the affine transformation determines the source coordinate $ \textbf{p}' = (x', y', z')$ as follows:

\begin{equation}
\small
 \begin{bmatrix}
    \mathbf{p}' \\
    1
 \end{bmatrix} =
\begin{bmatrix}
    1 & 0 & 0 & \theta_1 \\
    0 & 1 & 0 & \theta_2 \\
    0 & 0 & 1 & \theta_3 \\
    0 & 0 & 0 & 1
 \end{bmatrix}
\cdot
\begin{bmatrix}
    \mathbf{p} \\
    1
 \end{bmatrix}
\label{eq:transformation}
\end{equation}

Note that we only learn the shift elements of the affine matrix. 
Given the newly mapped coordinates, $Pr$ deforms to $\widehat{Pr}_{\text{affine}}$ with tri-linear resampling.
Once the objects are approximately aligned, the majority of the heavy deformation work is handled by the TPS deformation block.

\subsubsection{TPS block}
\label{sec:tps-block}
The TPS deformation block further deforms $\widehat{Pr}_{\text{affine}}$, producing $\widehat{Pr}$ as the final deformed output. 
TPS allows non-linear transforms using a set of source control vectors $\{\mathbf{p}_{\text{control}}^{(i)} \in \mathbb{R}^3 \}_{i=1}^{N}$, shown as the red arrows in Fig~\ref{fig:model}(b). We set $N = H_z * W_z * D_z$.
The control vectors are estimated by applying a convolution with 3 output channels.
Given the control vectors $\{\mathbf{p}_{\text{control}}^{(i)}\}_{i=1}^{N}$, TPS maps a target coordinate $\mathbf{p}$ to a source coordinate $\mathbf{p}'$ with

\begin{equation}
\label{eq:tps_function}
\small
\mathbf{p}' = \mathbf{A} \mathbf{p} + \sum_{i=1}^N U(|\mathbf{p} - \mathbf{p}_{\text{control}}^{(i)}|) * \mathbf{a}^{(i)}
\end{equation}
where $U(\mathbf{x}) = \mathbf{x}^2 \log \mathbf{x}^2$, $\mathbf{A}$ is a $3 \times 3$ matrix of linear coefficients, and $\mathbf{a}^{(i)}$ is a 3D vector of radial basis function coefficient.
The coefficients $\mathbf{A}$ and $\mathbf{a}^{(i)}$ are found by solving a linear equation with several constraints. Identical to affine block, resampling is done based on newly mapped source coordinates, producing $\widehat{\mathbf{Pr}}$ as the final output in deform block. Please refer to the supplementary material for more details. 



\subsection{Learnable Anatomical Prior}
Utilizing an accurate organ anatomy as a global prior significantly enhances the precision of the later adjusted global and local priors. 
Other atlas-based segmentation methods~\cite{kong2021deep,bongratz2023abdominal}, which assign a ground truth anatomy from a training set, are suboptimal.
Often, these scans do not cover the entire anatomy but only a small portion of it, making it impossible to recover an organ that does not exist in the chosen template.

AIC-Net learns to find the optimal global prior during training.
This is achieved by turning the global prior $\mathbf{Pr}_{g} \in \mathbb{R}^{C \times H_g\times W_g\times D_g}$ as a free parameter that needs to be optimized to produce an accurate prediction after a deformation.

Our experiments show that optimizing both the prior and other modules in AIC-Net leads to slower convergence. 
We hypothesize that this is due to a correlation between the prior and deformation modules.
For instance, if the predicted anatomy is smaller than the ground truth, the error can be reduced in two ways: 1) shrinking the source control points in TPS deformation or 2) enlarging the global prior. 
This correlation may confuse optimization priority. 
We prevent confusion by alternating the optimization of the model parameters and global prior.

\begin{figure}[!t]
    \centering
    \includegraphics[width=0.97\columnwidth]{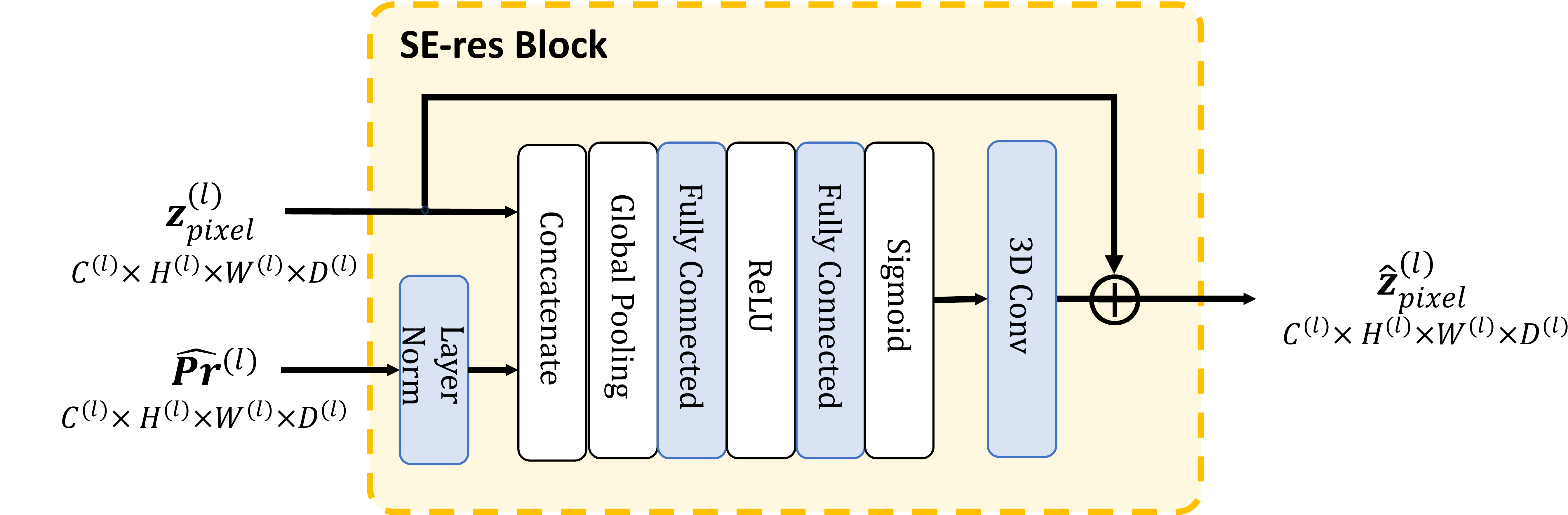}
    \caption{SE-res block is the Squeeze-and-Excitation block with a skip-connection which merges a decoder embedding $\mathbf{z}_{\text{decoder}}^{(l)}$ with a down-sized deformed prior $\widehat{\mathbf{Pr}}^{(l)}$ to produce a refined decoder embedding $\widehat{\mathbf{z}}_{\text{decoder}}^{(l)}$. Layer Norm normalizes high values in  $\widehat{\mathbf{Pr}}^{(l)}$.}
    \label{fig:se-res}
\end{figure}

\subsection{Aggregation of Anatomical Prior}
Fig~\ref{fig:se-res} illustrates SE-res block that merges a decoder embedding $\mathbf{z}_{\text{decoder}}^{(l)}$ and down-sized deformed prior $\widehat{\mathbf{Pr}}^{(l)}$ to produce a refined decoder embedding $\widehat{\mathbf{z}}_{\text{decoder}}^{(l)}$ at $l$'th decoder layer.
Layer normalization is employed on $\widehat{\mathbf{Pr}}^{(l)}$ to constrain its range, as it has been observed that the values of the learnable prior frequently fall within the interval of (-10, 10). 
Subsequently, we apply the Squeeze-and-Excitation~\cite{hu2018squeeze} operation to the concatenated features, followed by a convolutional layer to adjust the channel size to $C^{(i)}$. Additionally, a residual connection is introduced, allowing the module to bypass the computation if needed.


%

\subsection{Loss Function}

AIC-Net is trained to minimize 2 types of losses : Soft-Dice loss and centroid loss, and a regularizer. The first loss type is a Soft-Dice loss which measures the extent of overlap between the predicted and ground-truth mask. The second type of loss we introduce is centroid loss, which evaluates the alignment of the center of mass between the predicted and ground-truth organs. Regularizer is used to penalize deformation of prior that is too wild.

\subsubsection{Dice loss}

Given a prediction~$\widehat{\mathbf{Y}} \in [0,1]^{C_\text{cls} \times N}$ and its ground-truth~${\mathbf{Y}} \in {\{0,1\}}^{C_\text{cls} \times N}$, where $N = H \times W \times D$,  Soft-Dice is defined as 

\begin{equation}
\small
\mathbf{L}_{\text{dice}} (\mathbf{Y},\widehat{\mathbf{Y}}) = 1 - \frac{1}{C_\text{cls}} \sum_{c} \frac{2 \cdot \sum_{n} \widehat{y}_{c,n} \cdot y_{c,n}} {\sum_n  \widehat{y}_{c,n} + \sum_n y_{c,n} + \epsilon} 
\label{eq:placeholder}
\end{equation}
with $\epsilon$ at the denominator to handle the case when both ground-truth and predicted mask are empty.

\subsubsection{Centroid loss}

We have to be careful when applying dice loss on deformed priors $\widehat{\mathbf{Pr}}_g$ and $\widehat{\mathbf{Pr}}_l$.
In pixel-wise prediction, the Dice loss landscape with respect to the weights in the decoder block is typically not flat, making it favorable for gradient-based optimization.
However, in deformation-based prediction, the loss landscape with respect to the deformation parameter is flat in regions where there is no overlap between the deformed prior and the ground truth.
This occurs because the deformation parameters within the deform block solely dictate the amount of deformation.
Minor changes in these parameters lead to only local perturbations in the prior, which are insufficient to cause any overlap with the ground truth, thus causing the loss to remain unchanged.

To prevent the loss landscape from entering flat regions, it is essential to ensure some overlap between the ground truth and the deformed prior. We achieve this by introducing a novel centroid loss. Given an affine-transformed prior $\widehat{\mathbf{Pr}}_{\text{affine}}$ (shown in Fig~\ref{fig:model}(b)) and the ground truth $\mathbf{Y}$, the centroid loss is defined as the per-class averaged $L_2$ loss between the centroids of $\widehat{\mathbf{Pr}}_{\text{affine}}$ and $\mathbf{Y}$, i.e., $\{ \bar{\mathbf{g}}^{(c)}_{\text{Pr}} \}_{c = 1}^{C_{\text{cls}}}$ and $\{ \bar{\mathbf{g}}^{(c)}_{\mathbf{Y}} \}_{c = 1}^{C_{\text{cls}}}$:

\begin{equation}
\label{eq:centroid_loss}
\small
\mathbf{L}_{\text{centroid}} (\widehat{\mathbf{Pr}}_{\text{affine}}, \mathbf{Y}) = \frac{1}{C_\text{cls}}\sum_{c} \left\| \bar{\mathbf{g}}^{(c)}_{\text{Pr}} - \bar{\mathbf{g}}^{(c)}_{\mathbf{Y}} \right\|_2
\end{equation}
The centroid of each organ is defined as a simple object-wise spatial average. For example, the centroid of $c$'th organ from $\mathbf{Y}$ is computed as:
%
\begin{equation}
\small
\bar{\mathbf{g}}_{\mathbf{Y}}^{(c)} = \frac{1}{N_c(\textbf{Y})} \sum_{(h,w,d)} (h,w,d) \cdot \mathbb{I}_{\{(\text{argmax}_\text{channel} [ Y_{h,w,d} ]) = c\}}
\end{equation}
where $N_c(\textbf{Y})$ is the number of elements in $\mathbf{Y}$  that belong to organ $c$, and $h,w,d$ are grid coordinates.

\subsubsection{Final loss}

Combining the two loss function types, both at global and local level, as well as adding $L_2$ regularization terms gives the final loss function used to train AIC-Net:

\begin{align}
\label{eq:main_loss}
\small
\mathbf{L}_{\text{total}} = \sum_{v \in \{l,g\}} \left[ 
\begin{aligned}
&\mathbf{L}_{\text{dice}} (\mathbf{Y}_{v}, \widehat{\mathbf{Y}}_{v}) + 
  \mathbf{L}_{\text{dice}} (\mathbf{Y}_{v}, \widehat{\mathbf{Pr}}_{v}) + \\
& \gamma_{v}\mathbf{L}_{\text{centroid}} (\mathbf{Y}_{v}, \widehat{\mathbf{Pr}}_{\text{affine},v}) + \\
& \lambda_{v}  \sum_i \| \mathbf{p}_{\text{control},v}^{(i)} \|_2 
\end{aligned}
\right]
\end{align}
where $\mathbf{Y}_{v}$ is ground-truth labels at a view $v$, $\{\mathbf{p}_{\text{control}, v}^{(i)}\}$ are TPS source control points, and $\gamma_{v}$ and $\lambda_{v}$ are regularization hyper-parameters. We sum over global and local views.





\section{Experimental Detail}
\label{sec:experimental}

\subsection{Dataset}
To evaluate model performance, we use the publicly available TotalSegmentator Dataset~\cite{wasserthal2023totalsegmentator}. 
TotalSegmentator is a comprehensive dataset consisting of 1204 CT scans, divided into a training dataset of 1082 patients (90\%), a validation dataset of 57 patients (5\%), and a test dataset of 65 patients (5\%).
The dataset contains a wide variety of CT images, with differences in slice thickness, resolution, and scanning devices. 
The dataset also includes patients with abnomalities (tumor, bleeding and etc).
The dataset has 104 anatomic structures, which are sub-grouped into categories. We select the vertebrae and abdominal organs subgroups, which comprises 26 and 21 structures respectively.

The pixel intensity is truncated to the range \([-250, 1100]\) for vertebrae segmentation task, and to \([-250, 500]\) for organ segmentation. We normalize the spacing to \([1.5, 1.5, 2.0]\). The axial direction (\(d\)-dimension) is zero-padded to achieve a uniform volume size of $[288,288,512]$. Both the global input volume and the global mask are down-sampled by a factor of \([3, 3, 2]\). The dimensions of the local cropped views are set to \([128, 128, 128]\). Final predictions for evaluations are obtained by sliding window method on high resolution volumes.

\subsection{Training}

We use AdamW~\cite{loshchilov2017decoupled} optimizer with linear warmup cosine annealing. Maximum learning rate and weight-decay are set to $3\mathrm{e}{-4}$ and $1\mathrm{e}{-5}$. For the optimization of the prior, the learning rate is set to $1\mathrm{e}{-3}$.
Every 500 iterations, we conduct training for the prior over a span of 100 iterations.
The model is trained for 200K iterations. Batch size is set to 2. In the loss \eqref{eq:main_loss}, we set both $\lambda_{g}$ and $\lambda_{l}$ $1\mathrm{e}{-5}$. We set both $\gamma_{g}$ and $\gamma_{l}$ to $0.5$.




\begin{table*}[t!]
\caption{Comparison of AIC-Net and baseline on Organ and Vertebrae tasks with different backbones. The Vanilla model includes only the backbone segmentation network. The Cascaded model is a global-local approach similar to AIC-Net but does not have the learnable prior and the Deform block. \textbf{Best-performing} instances are in bold, while \underline{second-bests} are underlined. } 
\centering 
\begin{adjustbox}{width=0.85\textwidth}
\small
    \begin{tabular}{lcc ccc ccc }
        \toprule        
        {} & {} & {} & \multicolumn{3}{c}{\bfseries Organ} & \multicolumn{3}{c}{\bfseries Vertebrae} \\
        \cmidrule(l){4-6} \cmidrule(l){7-9} \\
        \bfseries{Model type} & \bfseries{Backbone} & \bfseries{Architecture} & \bfseries $\text{HD}_{95}$ {$\downarrow$} & \bfseries DSC {$\uparrow$} & \bfseries NSD {$\uparrow$} & \bfseries $\text{HD}_{95}$ {$\downarrow$} & \bfseries DSC {$\uparrow$} & \bfseries NSD {$\uparrow$} \\
        \cmidrule(l){1-3} \cmidrule(l){4-6} \cmidrule(l){7-9} \\
        
        {Vanilla} & {UNet} & {Convolutional} & {7.66} &{83.8} &{79.0}  & {11.3} &{85.6} &{73.8}   \\ 
        {Cascaded} & {UNet} & {Convolutional} & {\underline{6.46}} &{\underline{83.6}} &{\textbf{80.9}}  & {\underline{2.13}} &{\textbf{86.5}} &{\underline{93.7}}   \\ 
        {AIC-Net} & {UNet} & {Convolutional} & {\textbf{6.39}} &{\textbf{84.1}} &{\underline{80.4}}  & {\textbf{1.90}} &{\underline{86.2}} &{\textbf{94.0}}   \\ 

        \midrule

        {Vanilla} & {DeepLabV3+} & {Convolutional} & {7.56} &{\underline{79.7}} &{81.2}  & {\textbf{1.94}} &{\underline{82.9}} &{\textbf{94.1}}   \\ 
        {Cascaded} & {DeepLabV3+} & {Convolutional} & {\underline{7.28}} &{78.5} &{\underline{82.4}}  & {2.06} &{82.6} &{93.6}   \\ 
        {AIC-Net} & {DeepLabV3+} & {Convolutional} & \textbf{4.28} &{\textbf{80.2}} &{\textbf{84.5}}  & {\textbf{1.94}} &{\textbf{83.2}} &{\textbf{94.1}}   \\ 
        
        \midrule

        {Vanilla} & {UNETR} & {Transformer} & {27.1} & {71.1} &{59.4}  & {52.2} &{\underline{76.5}} &{53.2}   \\ 
        {Cascaded} & {UNETR} & {Transformer} & {\textbf{12.0}} &{\underline{75.2}} & {\textbf{70.7}}  & {\underline{15.6}} &{76.2} &{\underline{64.1}}   \\ 
        {AIC-Net} & {UNETR} & {Transformer} & {\underline{14.4}} &{\textbf{75.4}} &{\underline{69.1}}  & {\textbf{12.6}} &{\textbf{83.2}} &{\textbf{74.4}}   \\

        \midrule

        {Vanilla} & {UNETR-Swin} & {Hybrid} & {7.89} &{\underline{84.1}} &{76.7}  & {\underline{6.59}} &{\textbf{90.2}} &{\underline{79.2}}   \\ 
        {Cascaded} & {UNETR-Swin} & {Hybrid} & {\underline{6.42}} &{83.8} &{\underline{79.0}}  & {12.6} &{88.5} &{76.0}   \\ 
        {AIC-Net} & {UNETR-Swin} & {Hybrid} & {\textbf{6.18}} &{\textbf{84.2}} &{\textbf{80.4}}  & {\textbf{1.76}} &{\underline{89.3}} &{\textbf{95.3}}  \\ 
        \bottomrule
    \end{tabular}
\end{adjustbox}
\label{table:accruacy}
\end{table*}

\section{Results and Discussion}

\subsection{Segmentation Performance}

We evaluate AIC-Net with four widely used backbones for medical image segmentation tasks: UNet~\cite{ronneberger2015u}, DeepLabV3+~\cite{chen2018encoder}, UNETR~\cite{hatamizadeh2022unetr}, and UNETR-Swin~\cite{hatamizadeh2021swin}.
For each backbone and segmentation task, we evaluate three model types: Vanilla, Cascaded, and AIC-Net. The Vanilla model includes only the backbone segmentation network. The Cascaded model is a global-local approach similar to AIC-Net but does not incor-
porate the learnable prior and the Deform block.
We measure segmentation performance by three metrics: the dice score (DSC), the normalized surface dice (NSD), and the $95\%$ Hausdorff distance ($\text{HD}_{95}$).

As shown in Table~\ref{table:accruacy}, AIC-Net consistently outperforms the other two baselines across all three metrics.
Notably, the performance improvement is more pronounced in terms of $\text{HD}_{95}$ compared to the other two metrics.
HD is a superior metric for assessing the anatomical accuracy of predictions.
Unlike DSC and NSD, which evaluate the extent of overlap, HD measures the maximum pixel difference, making it a more precise indicator of anatomical correctness, as it significantly penalizes mis-predictions that are distant from the ground truth.

\subsection{Impact of Centroid Loss}

\begin{figure}[htbp]
     \centering
     \begin{subfigure}[b]{0.59\columnwidth}
         \centering
         \includegraphics[width=\textwidth]{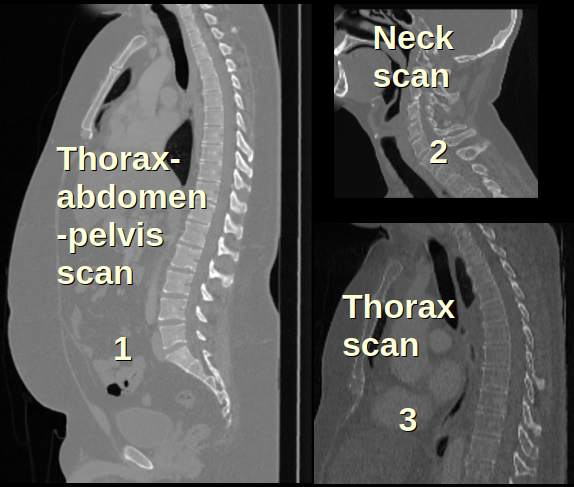}
         \caption{Common scan types}
         \label{fig:CT_3_types}
     \end{subfigure}
     \hfill
     \begin{subfigure}[b]{0.19\columnwidth}
         \centering
         \includegraphics[width=\textwidth]{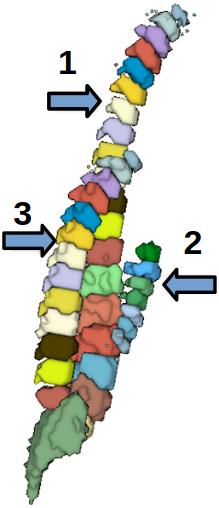}
         \caption{without \\ centroid loss}
         \label{fig:prior_no_centroid_loss}
     \end{subfigure}
     \hfill
     \begin{subfigure}[b]{0.19\columnwidth}
         \centering
         \includegraphics[width=\textwidth]{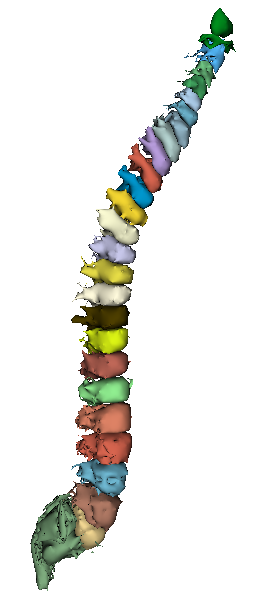}        
         \caption{with \\ centroid loss}
         \label{fig:prior_with_centroid_loss}
     \end{subfigure}
     \hfill
    \caption{Impact of centroid loss. A common vertebrae configuration should be learned, while the Deform Block align the prior to right positions. (a) Three common scan types in dataset. Scans always appear at center of padded volume. (b) Without centroid loss (failed case): Deform Block fails to shift with large displacement, and learned prior are forced to adopt three vertebrae configurations. (c) With centroid loss, we can learn a prior with correct anatomy.}
        \label{fig:impact_of_centroid_loss}
\end{figure}

The centroid loss (CL) introduced in \eqref{eq:centroid_loss} is essential for learning common prior and robust deformations. Figure~\ref{fig:prior_no_centroid_loss} shows the learned prior without CL, which resulted in three sets of vertebrae configurations as indicated by the arrows. These corresponds to the three common scanning positions in the dataset (thorax-abdomen-pelvis, neck, and thorax scans) as shown in Figure~\ref{fig:CT_3_types}. Without CL, the Deform block fails to properly shift the prior to correct positions, and the learnable prior is forced to represent three vertebrae configurations. With CL, deformation prior can successfully align a unique set of vertebrae configuration to all types of scans, resulting in a much better prior. 

\begin{table}[t!]
\caption{Impact of Deform block on deformed local prior accuracy. The \textbf{best-performing} instance is highlighted in bold.} 
\centering 
\begin{adjustbox}{width=1\columnwidth}
\small
    \begin{tabular}{lc cc cc }
        \toprule
        {} & {} & \multicolumn{2}{c}{\bfseries Organ} & \multicolumn{2}{c}{\bfseries Vertebrae} \\ 
        \cmidrule(l){3-4} \cmidrule(l){5-6} \\
        \bfseries Deform & \bfseries Backbone & \bfseries $\text{HD}_{95}$ {$\downarrow$} & \bfseries DSC {$\uparrow$}  &\bfseries $\text{HD}_{95}$ {$\downarrow$} & \bfseries DSC {$\uparrow$} \\
        \cmidrule(l){3-4} \cmidrule(l){5-6} \\
        {no} & {UNet} & \textbf{7.26} & 60.8   & 4.92 & 55.4 \\
        {yes} & {UNet} & 7.45 & \textbf{67.6}   & \textbf{3.70} & \textbf{68.9} \\
        \midrule
        {no} & {DeepLabV3+} & 8.43 &  57.5  & 5.93 & 53.4 \\
        {yes} & {DeepLabV3+} & \textbf{6.37} & \textbf{68.1}   & \textbf{4.24} & \textbf{64.2} \\
        \midrule
        {no} & {UNETR} & 13.8 & 57.2   & 7.35 & 45.5 \\
        {yes} & {UNETR} & \textbf{11.7} & \textbf{65.9}   & \textbf{6.45} & \textbf{58.1} \\
        \midrule
        {no} & {UNETR-Swin} & 7.27 & 60.6   & 5.49 & 51.2 \\
        {yes} & {UNETR-Swin} & \textbf{6.64} & \textbf{70.7}   & \textbf{3.99} & \textbf{71.0} \\        
        \bottomrule
    \end{tabular}
\end{adjustbox}
\label{table:deform}
\end{table}

\subsection{Deformation Performance}
We assess the impact of the Deform block by comparing the accuracy of the prior at the local level before deformation $\mathbf{Pr}_l$ and after deformation $\widehat{\mathbf{Pr}}_l$. Table~\ref{table:deform} demonstrates that our Deform block refines a coarse prior into a more fine-grained one.

\subsection{Visualization of Deformed Prior}
\begin{figure}[ htbp]
     \centering
     \includegraphics[width=0.8\columnwidth]{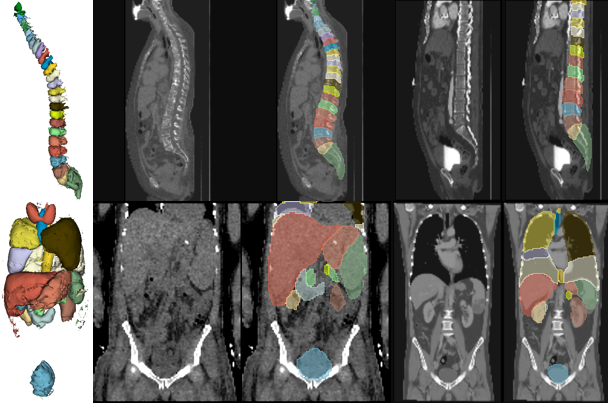}
    \caption{Visualizations of learned common priors (left) and their deformation to patient-specific anatomies (right).}
        \label{fig:deformation_vis}
\end{figure}

Figure~\ref{fig:deformation_vis} shows the learned global priors and patient-specific deformed anatomy $\widehat{\mathbf{Pr}}_g$ learned by the TPS deform block as illustrated in Figure~\ref{fig:model}(b). The figure depicts that the learned global priors closely align with our understanding of generic anatomies of vertebrae and abdominal organs. Additionally, the prior anatomy is successfully deformed into different patient-specific anatomies. For instance, the spine anatomy in the left scan shows greater curvature, while the spine anatomy in the right scan appears straighter; the overall positions of the vertebrae also differ significantly.

\begin{figure}[htbp]
     \centering
     \includegraphics[width=0.9\columnwidth]{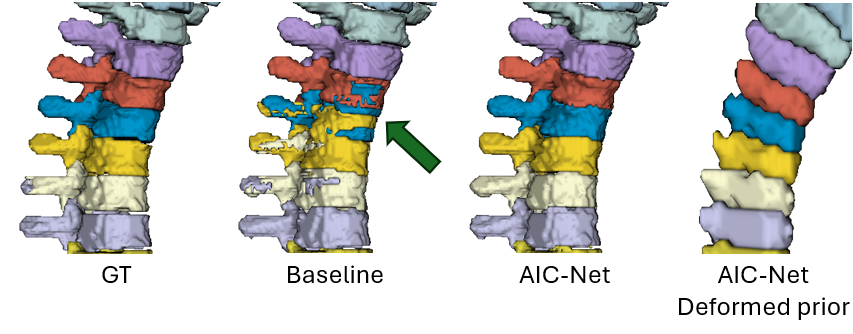}
    \caption{Qualitative comparisons on vertebrae segmentation. Baseline model produces mixed labels, as indicated by the green arrow. For AIC-Net, since the deformed prior already gives good indications of relative positions of vertebrae, it facilitates the identification of each vertebra in the final prediction.}
    \label{fig:vertebrae_comparison}
\end{figure}

\begin{figure}[htbp]
     \centering
         \includegraphics[width=0.85\columnwidth]{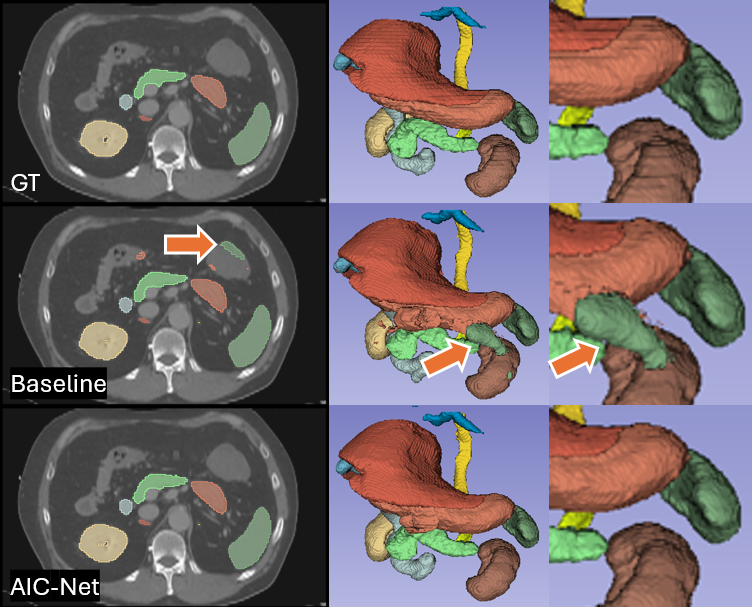}
    \caption{Qualitative comparisons on organ segmentation. The baseline method is suboptimal, resulting in the segmentation of additional spleen tissue, as indicated by the orange arrows.}
        \label{fig:organ_comparison_308}
\end{figure}

\begin{figure}[htbp]
     \centering
         \includegraphics[width=0.85\columnwidth]{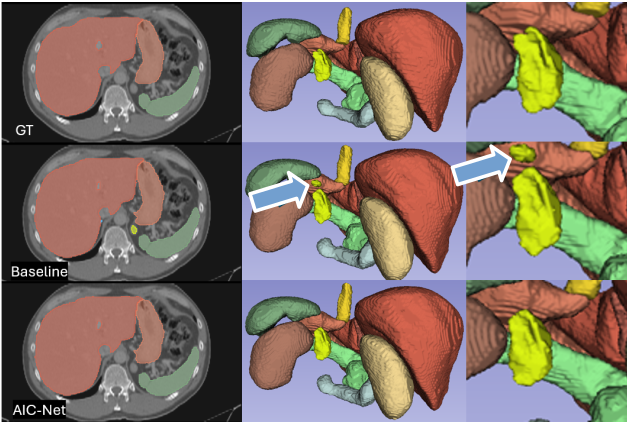}
    \caption{Qualitative comparisons on organ segmentation. The baseline method (actually all baseline backbones) incorrectly segments the adrenal gland, as shown by the blue arrows.}
        \label{fig:organ_comparison_311}
\end{figure}

\subsubsection{Qualitative Comparison}
The learned common prior, as well as accurate deformation, in our AIC-Net can promote anatomically accurate segmentation. This is supported by results in Figures~\ref{fig:vertebrae_comparison}, \ref{fig:organ_comparison_308} and \ref{fig:organ_comparison_311}. In Figure~\ref{fig:vertebrae_comparison}, despite being over-smoothed, the deformed prior at the global level still provides accurate guidance for identifying vertebra indices, which in turn supports precise segmentation at the local level. In contrast, the baseline method appears to struggle with correctly identifying vertebra indices, leading to inconsistent predictions. We also observe that this mixing effect is a common issue in bone segmentation tasks \cite{wasserthal2023totalsegmentator}. In Figures~\ref{fig:organ_comparison_308} and \ref{fig:organ_comparison_311}, baseline methods give incorrect segmentation that result in separated spleen and left adrenal gland, which clearly violate human anatomy. For both cases, AIC-Net gives correct predictions.

\section{Conclusion}
AIC-Net is a general approach that enhances existing segmentation models by incorporating a learnable anatomical prior, which adapts to patient-specific anatomy using differentiable spatial deformation functions, making the models more anatomy-aware.

Though AIC-Net offers a performance boost, it also has several drawbacks and potential rooms for improvement.
AIC-Net is not a cheap model. It is a global-local cascaded model that doubles model size and memory consumption. Thus a potential future research direction could focus on learning the prior without the guidance from the global view.
Also, though AIC-Net does not use the deformed-prior as the final prediction but as a soft constraints to the decoder output, it is still desirable for the deformed-prior to have more accurate fine-grained predicitons as this will ease the fusion between the deformed prior and decoder outputs. Thus future research could consider replacing the deformation functions in the Deform block to a more flexible one that is better at representing fine-grained objects.

\newpage

\cleardoublepage

\section{Thin-plate-spline Deformation}
In this section, we provide a detailed explanation of thin-plate splines (TPS)~\cite{bookstein1989principal}, including its unraveled non-matrix definition and the optimization of TPS coefficients by solving a linear equation subject to several constraints. As per convention in computer vision, we call the three coordinate axes in $\mathbb{R}^3$ the $h$, $w$ and $d$-axis.


Given an object $P$ in $\mathbb{R}^3$, we wish to alter its shape by warping the coordinate axes. This can be done by constructing a warping function $\mathcal D: \mathbb{R}^3 \to \mathbb{R}^3$, and reconstruct the new shape $Y$ by
\begin{equation*}
    (h',w',d') = \mathcal{D}(h,w,d), \quad Y(h,w, d) := P(h',w',d').
\end{equation*}
That is, the value of the \textit{target} object $Y$ at the coordinate point $(h,w,d)$, is given by the value of the \textit{source} object $P$ at the warped coordinate point $(h',w',d')$ which is calculated by the warping function $\mathcal{D}$. We call points associated with the target object $Y$ \textit{target points}, and points associated with the source object $P$ \textit{source points}.

The thin-plate-spline deformation is a method to construct the warping function $\mathcal{D}$. Given a sequence of \textit{target control points} $\{\mathbf{p}_i \}_i^N$ and a corresponding \textit{source control points} $\{ \mathbf{p}'_i \}_i^N$, the warping $\mathcal{D}$ maps exactly $\mathbf{p}_i\mapsto \mathbf{p}'_i$ with minimal bending energy. Given a general target point $\mathbf{p}=(h,w,d)$, its image under $\mathcal{D}$ is given by
\begin{subequations}
    \label{eq:tps_xy_functions}
\begin{align}
    \mathcal{D}_h(\mathbf{p}) &= a^{(N+1)} + a^{(N+2)} h + a^{(N+3)} w + a^{(N+4)} d \nonumber \\
    &\quad + \sum_{i=1}^N a^{(i)} U(|\mathbf{p} - \mathbf{p}_i|), \label{eq:tps_h_function} \\
    \mathcal{D}_w(\mathbf{p}) &= b^{(N+1)} + b^{(N+2)} h + b^{(N+3)} w + b^{(N+4)} d \nonumber \\
    &\quad + \sum_{i=1}^N b^{(i)} U(|\mathbf{p} - \mathbf{p}_i|), \label{eq:tps_w_function} \\
    \mathcal{D}_d(\mathbf{p}) &= c^{(N+1)} + c^{(N+2)} h + c^{(N+3)} w + c^{(N+4)} d \nonumber \\
    &\quad + \sum_{i=1}^N c^{(i)} U(|\mathbf{p} - \mathbf{p}_i|), \label{eq:tps_d_function}
\end{align}
\end{subequations}
where $U(r) = r^2\log r^2$ is the kernel function, $(a^{(1)},\cdots, a^{(N+4)})$, $(b^{(1)},\cdots, b^{(N+4)})$, and $(c^{(1)},\cdots, c^{(N+4)})$ are TPS coefficients that are determined by mapping the control points.  The TPS coefficients can be obtained by solving a linear equation.

Here, we use the $h$-coordinate coefficients as an example, and the calculation of the $w$ and $d$ coordinate coefficients are done in a similar manner. The function~\eqref{eq:tps_h_function} has $N+4$ coefficients to be computed, which can be calculated by a closed-form solution.

Let $\mathbf{v} = (h'_1, \cdots, h'_N| 0,0,0,0)^T$, where $h'_i$ is the $h$-coordinate of the $i$-th source control point. Also, define matrices
\begin{subequations}
    \label{eq::appendix_tps0}
\begin{align}
    \mathcal{K} = &
    \begin{bmatrix}
        0 & U_{12} & \cdots & U_{1N} \\
        U_{21} & 0 & \cdots & U_{2N} \\
        \cdots & \cdots & \cdots & \cdots \\
        U_{N1} & U_{N2} & \cdots & 0 \\    
    \end{bmatrix}
    , N \times N; \nonumber\\ 
    \mathcal{P} = &
    \begin{bmatrix}
        1 & h_1 & w_1 & d_1\\
        1 & h_2 & w_2 & d_2\\
        \cdots & \cdots & \cdots & \cdots \\
        1 & h_N & w_N & d_N\\
    \end{bmatrix} 
    ,N \times 4; \nonumber\\ 
    \mathcal{M} = &
    \begin{bmatrix}
        \mathcal{K} & \mathcal{P} \\
        \mathcal{P}^T & O \\
    \end{bmatrix}
    ,(N+4) \times (N+4)
\end{align}
\end{subequations}
where $U_{i,j} = U(|\mathbf{p}_i- \mathbf{p}_j|)$, $h_i$, $w_i$, and $d_i$ are the $h$-,  $w$-, and $d$-coordinates of the target control point $\mathbf{p}_i$, and $O$ is a zero matrix of size $4 \times 4$. Then the coefficients $ \mathbf{a} =(a^{(1)},\cdots, a^{(N+4)})$ are given by
\begin{equation}\label{eq::appendix_tps1}
\mathbf{a} = \mathcal{M}^{-1}\mathbf{v}.
\end{equation}
The additional last four rows of $\mathcal{M}$ guarantee that the coefficients $a^{(i)}$ sum to zero and that their cross-products with the points $\mathbf{p}_i$ are likewise zero. These extra conditions are regularization terms used in TPS formulation.

In our implementation, we keep the target control points fixed, and use neural networks to propose the source control points. By doing so, we only need to calculate $\mathcal{M}^{-1}$ once, and we do not have numerical instability problem.

\section{Extended Results}

We present the class-wise performance of all backbone models for both the Organ and Vertebrae tasks in the following tables. Notably, AIC-Net not only achieves better mean scores but also generally exhibits lower standard deviations, indicating more consistent and reliable performance. We have omitted the segmentation results for kidney\_cyst\_left and kidney\_cyst\_right from the Organ task, as both AIC-Net and the baseline models failed to predict these classes.

\begin{table*}[]
\caption{Organ Segmentation Comparison on UNet Backbone}
\centering
\label{tab:organ_comparison_unet_swin}
\begin{adjustbox}{width=0.8\textwidth}
\small
 
\begin{tabular}{rcccccccc}
{} & \multicolumn{2}{c}{\bfseries{NSD} $\uparrow$} & & \multicolumn{2}{c}{\bfseries{$\text{Haus}_{95} \downarrow $}} & & \multicolumn{2}{c}{\bfseries{Dice} $\uparrow$} \\
\cmidrule{2-3} \cmidrule{5-6} \cmidrule{8-9}
& AIC-Net & Baseline & & AIC-Net & Baseline & & AIC-Net & Baseline \\
adrenal\_gland\_left & 91.9 $\pm$ 22.7 & 94.0 $\pm$ 14.5 & & 1.94 $\pm$ 1.79 & 2.59 $\pm$ 3.27 & & 79.9 $\pm$ 22.1 & 80.5 $\pm$ 16.9 \\
adrenal\_gland\_right & 97.3 $\pm$ 9.00 & 97.3 $\pm$ 8.92 & & 1.55 $\pm$ 1.31 & 1.66 $\pm$ 1.49 & & 83.5 $\pm$ 13.2 & 82.3 $\pm$ 13.8 \\
colon & 85.6 $\pm$ 19.0 & 86.6 $\pm$ 16.2 & & 11.4 $\pm$ 16.1 & 11.1 $\pm$ 15.1 & & 86.2 $\pm$ 12.0 & 85.2 $\pm$ 13.9 \\
duodenum & 88.7 $\pm$ 17.3 & 87.5 $\pm$ 18.1 & & 3.40 $\pm$ 2.59 & 3.98 $\pm$ 3.48 & & 80.7 $\pm$ 18.9 & 79.0 $\pm$ 19.4 \\
esophagus & 97.6 $\pm$ 5.32 & 94.7 $\pm$ 16.9 & & 2.70 $\pm$ 5.56 & 5.37 $\pm$ 24.2 & & 89.1 $\pm$ 5.30 & 88.6 $\pm$ 5.15 \\
gallbladder & 82.4 $\pm$ 31.9 & 79.1 $\pm$ 34.4 & & 3.88 $\pm$ 5.06 & 5.30 $\pm$ 7.68 & & 77.4 $\pm$ 30.9 & 79.7 $\pm$ 26.2 \\
kidney\_left & 91.1 $\pm$ 22.2 & 94.1 $\pm$ 15.4 & & 5.89 $\pm$ 15.3 & 3.72 $\pm$ 8.21 & & 91.0 $\pm$ 16.4 & 91.1 $\pm$ 16.0 \\
kidney\_right & 88.7 $\pm$ 27.6 & 87.7 $\pm$ 28.9 & & 3.77 $\pm$ 7.50 & 3.62 $\pm$ 6.87 & & 91.9 $\pm$ 16.4 & 92.2 $\pm$ 14.6 \\
liver & 91.6 $\pm$ 20.5 & 89.5 $\pm$ 24.3 & & 4.77 $\pm$ 10.5 & 7.53 $\pm$ 20.1 & & 95.5 $\pm$ 8.15 & 95.0 $\pm$ 11.9 \\
lung\_lower\_lobe\_left & 90.9 $\pm$ 18.3 & 90.3 $\pm$ 20.6 & & 4.63 $\pm$ 14.8 & 6.72 $\pm$ 23.7 & & 91.7 $\pm$ 14.9 & 92.5 $\pm$ 13.1 \\
lung\_lower\_lobe\_right & 92.0 $\pm$ 16.8 & 93.1 $\pm$ 11.2 & & 2.81 $\pm$ 3.60 & 2.61 $\pm$ 2.47 & & 92.4 $\pm$ 13.7 & 93.1 $\pm$ 11.2 \\
lung\_middle\_lobe\_right & 86.1 $\pm$ 18.8 & 90.5 $\pm$ 9.74 & & 9.56 $\pm$ 27.8 & 3.92 $\pm$ 3.67 & & 89.8 $\pm$ 10.6 & 90.6 $\pm$ 8.79 \\
lung\_upper\_lobe\_left & 88.9 $\pm$ 22.7 & 90.3 $\pm$ 19.9 & & 4.04 $\pm$ 8.10 & 4.13 $\pm$ 9.55 & & 93.4 $\pm$ 6.92 & 93.4 $\pm$ 6.18 \\
lung\_upper\_lobe\_right & 71.1 $\pm$ 39.1 & 65.6 $\pm$ 42.9 & & 7.52 $\pm$ 15.9 & 12.8 $\pm$ 35.8 & & 86.3 $\pm$ 24.7 & 87.8 $\pm$ 22.8 \\
pancreas & 88.0 $\pm$ 23.7 & 89.0 $\pm$ 21.8 & & 2.98 $\pm$ 2.74 & 3.72 $\pm$ 4.62 & & 82.1 $\pm$ 22.9 & 83.1 $\pm$ 19.2 \\
prostate & 46.7 $\pm$ 44.9 & 43.7 $\pm$ 44.7 & & 2.97 $\pm$ 1.43 & 3.34 $\pm$ 2.39 & & 82.1 $\pm$ 12.2 & 81.4 $\pm$ 13.6 \\
small\_bowel & 85.9 $\pm$ 20.4 & 85.0 $\pm$ 20.5 & & 8.98 $\pm$ 14.8 & 15.6 $\pm$ 45.9 & & 86.4 $\pm$ 12.6 & 85.4 $\pm$ 13.0 \\
spleen & 94.1 $\pm$ 17.5 & 95.8 $\pm$ 12.8 & & 7.05 $\pm$ 27.7 & 9.27 $\pm$ 36.1 & & 96.0 $\pm$ 2.68 & 96.4 $\pm$ 1.64 \\
stomach & 88.6 $\pm$ 24.8 & 89.5 $\pm$ 21.7 & & 3.93 $\pm$ 6.75 & 4.71 $\pm$ 6.98 & & 90.2 $\pm$ 18.7 & 90.9 $\pm$ 14.7 \\
thyroid\_gland & 92.7 $\pm$ 21.8 & 75.6 $\pm$ 40.9 & & 8.41 $\pm$ 41.8 & 2.16 $\pm$ 2.32 & & 86.4 $\pm$ 8.60 & 85.2 $\pm$ 10.2 \\
trachea & 95.8 $\pm$ 15.5 & 84.4 $\pm$ 34.6 & & 5.24 $\pm$ 23.9 & 10.3 $\pm$ 41.6 & & 91.5 $\pm$ 5.37 & 91.1 $\pm$ 6.24 \\
urinary\_bladder & 81.3 $\pm$ 26.8 & 81.5 $\pm$ 26.1 & & 8.86 $\pm$ 20.9 & 19.9 $\pm$ 53.7 & & 87.3 $\pm$ 15.8 & 87.4 $\pm$ 15.4 \\
mean & 80.4 $\pm$ 22.0 & 79.0 $\pm$ 22.5 & & 6.39 $\pm$ 13.1 & 7.66 $\pm$ 16.9 & & 84.1 $\pm$ 16.2 & 83.8 $\pm$ 15.1
\end{tabular}
\end{adjustbox}
\end{table*}

\begin{table*}[]
\caption{Vertebrae Segmentation Comparison on UNet Backbone}
\centering
\label{tab:vertebra_comparison_unet}
\begin{adjustbox}{width=0.8\textwidth}
\small
 
    \begin{tabular}{rcccccccc}
    {}                        & \multicolumn{2}{c}{\bfseries{NSD} $\uparrow$}                      &  & \multicolumn{2}{c}{\bfseries{$\text{Haus}_{95} \downarrow $}}                    &  &  \multicolumn{2}{c}{\bfseries{Dice} $\uparrow$}                   \\
    \cmidrule{2-3} \cmidrule{5-6} \cmidrule{8-9}
                              & AIC-Net          & Baseline          &  & AIC-Net            & Baseline           &  & AIC-Net           & Baseline          \\
    sacrum                    & 84.2 $\pm$ 33.1  & 73.8 $\pm$ 40.3  &  & 2.17 $\pm$ 2.85   & 9.53 $\pm$ 27.7   &  & 88.2 $\pm$ 18.9   & 86.4 $\pm$ 20.6  \\
    vertebrae\_C1             & 91.7 $\pm$ 21.2  & 21.0 $\pm$ 38.1  &  & 2.63 $\pm$ 2.28   & 61.8 $\pm$ 83.6   &  & 78.4 $\pm$ 20.2   & 73.0 $\pm$ 20.1  \\
    vertebrae\_C2             & 96.0 $\pm$ 7.22   & 27.4 $\pm$ 42.9  &  & 3.08 $\pm$ 5.16   & 34.4 $\pm$ 87.9   &  & 81.5 $\pm$ 13.6   & 79.4 $\pm$ 15.0  \\
    vertebrae\_C3             & 99.3 $\pm$ 1.03   & 36.6 $\pm$ 48.5  &  & 1.22 $\pm$ 0.41   & 1.33 $\pm$ 0.56   &  & 86.0 $\pm$ 8.02    & 86.9 $\pm$ 4.07   \\
    vertebrae\_C4             & 92.1 $\pm$ 25.6  & 41.2 $\pm$ 48.9  &  & 1.39 $\pm$ 0.69   & 7.24 $\pm$ 21.4   &  & 79.6 $\pm$ 22.6   & 78.4 $\pm$ 22.7  \\
    vertebrae\_C5             & 87.5 $\pm$ 30.3  & 45.9 $\pm$ 48.4  &  & 1.55 $\pm$ 0.86   & 6.65 $\pm$ 20.6   &  & 73.4 $\pm$ 29.4   & 73.5 $\pm$ 23.3  \\
    vertebrae\_C6             & 82.2 $\pm$ 36.1  & 59.4 $\pm$ 46.7  &  & 1.54 $\pm$ 1.07   & 21.0 $\pm$ 54.0   &  & 68.8 $\pm$ 33.6   & 71.2 $\pm$ 26.2  \\
    vertebrae\_C7             & 99.1 $\pm$ 1.78   & 69.7 $\pm$ 44.8  &  & 1.27 $\pm$ 0.90   & 21.8 $\pm$ 58.5   &  & 90.7 $\pm$ 2.07    & 89.4 $\pm$ 2.15   \\
    vertebrae\_L1             & 95.6 $\pm$ 18.0  & 94.2 $\pm$ 20.9  &  & 1.43 $\pm$ 1.90   & 9.44 $\pm$ 26.7   &  & 91.4 $\pm$ 13.6   & 92.1 $\pm$ 13.3  \\
    vertebrae\_L2             & 97.4 $\pm$ 13.2  & 94.1 $\pm$ 20.9  &  & 1.51 $\pm$ 1.89   & 9.51 $\pm$ 30.1   &  & 91.9 $\pm$ 12.7   & 92.4 $\pm$ 11.9  \\
    vertebrae\_L3             & 95.8 $\pm$ 18.2  & 91.9 $\pm$ 24.8  &  & 1.45 $\pm$ 1.96   & 6.72 $\pm$ 23.8   &  & 93.8 $\pm$ 2.71    & 93.1 $\pm$ 5.78   \\
    vertebrae\_L4             & 97.2 $\pm$ 13.8  & 87.3 $\pm$ 30.4  &  & 1.45 $\pm$ 2.39   & 5.25 $\pm$ 24.6   &  & 91.7 $\pm$ 13.6   & 90.4 $\pm$ 13.6  \\
    vertebrae\_L5             & 99.1 $\pm$ 2.16   & 90.2 $\pm$ 27.2  &  & 1.23 $\pm$ 0.78   & 4.83 $\pm$ 24.0   &  & 93.4 $\pm$ 3.16    & 91.9 $\pm$ 4.69   \\
    vertebrae\_S1             & 96.8 $\pm$ 11.6  & 86.9 $\pm$ 30.8  &  & 1.74 $\pm$ 2.37   & 2.04 $\pm$ 2.10   &  & 90.0 $\pm$ 13.5   & 88.6 $\pm$ 12.4  \\
    vertebrae\_T1             & 99.5 $\pm$ 1.33   & 78.5 $\pm$ 40.7  &  & 1.24 $\pm$ 1.15   & 1.26 $\pm$ 1.04   &  & 92.3 $\pm$ 1.87    & 91.9 $\pm$ 1.66   \\
    vertebrae\_T10            & 93.8 $\pm$ 18.1  & 91.0 $\pm$ 23.4  &  & 2.37 $\pm$ 3.44   & 5.49 $\pm$ 19.5   &  & 86.5 $\pm$ 21.3   & 88.6 $\pm$ 16.4  \\
    vertebrae\_T11            & 93.8 $\pm$ 20.4  & 85.6 $\pm$ 31.9  &  & 1.94 $\pm$ 3.09   & 5.80 $\pm$ 20.9   &  & 88.8 $\pm$ 18.3   & 90.5 $\pm$ 10.2  \\
    vertebrae\_T12            & 95.5 $\pm$ 18.4  & 92.5 $\pm$ 22.8  &  & 2.24 $\pm$ 5.72   & 7.68 $\pm$ 34.0   &  & 90.2 $\pm$ 18.9   & 90.2 $\pm$ 16.6  \\
    vertebrae\_T2             & 98.8 $\pm$ 5.00  & 81.5 $\pm$ 37.6  &  & 1.29 $\pm$ 1.31   & 1.86 $\pm$ 2.00   &  & 91.3 $\pm$ 6.64   & 91.0 $\pm$ 6.48  \\
    vertebrae\_T3             & 97.4 $\pm$ 11.9  & 86.8 $\pm$ 30.2  &  & 1.80 $\pm$ 2.23   & 6.43 $\pm$ 20.6   &  & 89.4 $\pm$ 12.7   & 87.8 $\pm$ 14.2  \\
    vertebrae\_T4             & 92.9 $\pm$ 22.2  & 85.1 $\pm$ 32.4  &  & 2.47 $\pm$ 4.50   & 2.29 $\pm$ 2.68   &  & 86.2 $\pm$ 20.4   & 83.6 $\pm$ 23.4  \\
    vertebrae\_T5             & 91.8 $\pm$ 21.8  & 78.2 $\pm$ 37.9  &  & 2.41 $\pm$ 2.63   & 3.98 $\pm$ 6.13   &  & 83.7 $\pm$ 20.9   & 85.1 $\pm$ 18.4  \\
    vertebrae\_T6             & 93.2 $\pm$ 17.3  & 71.8 $\pm$ 40.9  &  & 2.32 $\pm$ 2.64   & 5.23 $\pm$ 13.5   &  & 83.4 $\pm$ 21.0   & 79.9 $\pm$ 23.7  \\
    vertebrae\_T7             & 88.1 $\pm$ 27.9  & 78.1 $\pm$ 35.6  &  & 3.13 $\pm$ 6.20   & 17.2 $\pm$ 44.7   &  & 79.5 $\pm$ 27.7   & 77.5 $\pm$ 27.5  \\
    vertebrae\_T8             & 90.7 $\pm$ 26.0  & 82.0 $\pm$ 33.9  &  & 2.63 $\pm$ 4.69   & 25.6 $\pm$ 58.0   &  & 82.9 $\pm$ 26.1   & 83.6 $\pm$ 22.1  \\
    vertebrae\_T9             & 94.9 $\pm$ 17.6  & 88.7 $\pm$ 27.3  &  & 1.98 $\pm$ 2.85   & 9.17 $\pm$ 29.0   &  & 87.4 $\pm$ 19.8   & 89.3 $\pm$ 14.2  \\
    mean                      & 94.0 $\pm$ 16.9  & 73.8 $\pm$ 34.9  &  & 1.90 $\pm$ 2.54   & 11.3 $\pm$ 28.4   &  & 86.2 $\pm$ 16.3   & 85.6 $\pm$ 15.0
    \end{tabular}
\end{adjustbox}
\end{table*}

\begin{table*}[]
\caption{Organ Segmentation Comparison on DeepLabV3+ Backbone}
\centering
\label{tab:organ_comparison_deeplab}
\begin{adjustbox}{width=0.8\textwidth}
\small
 
\begin{tabular}{rcccccccc}
{} & \multicolumn{2}{c}{\bfseries{NSD} $\uparrow$} & & \multicolumn{2}{c}{\bfseries{$\text{Haus}_{95} \downarrow $}} & & \multicolumn{2}{c}{\bfseries{Dice} $\uparrow$} \\
\cmidrule{2-3} \cmidrule{5-6} \cmidrule{8-9}
& AIC-Net & Baseline & & AIC-Net & Baseline & & AIC-Net & Baseline \\
adrenal\_gland\_left & 92.6 $\pm$ 19.4 & 90.3 $\pm$ 23.0 & & 2.18 $\pm$ 1.49 & 2.93 $\pm$ 3.04 & & 71.6 $\pm$ 19.7 & 69.7 $\pm$ 20.3 \\
adrenal\_gland\_right & 95.7 $\pm$ 10.3 & 94.5 $\pm$ 14.8 & & 2.14 $\pm$ 2.03 & 2.32 $\pm$ 2.55 & & 74.1 $\pm$ 15.6 & 71.8 $\pm$ 17.2 \\
colon & 88.6 $\pm$ 18.3 & 85.4 $\pm$ 21.6 & & 7.20 $\pm$ 10.3 & 8.87 $\pm$ 12.1 & & 87.6 $\pm$ 14.5 & 86.2 $\pm$ 15.5 \\
duodenum & 87.6 $\pm$ 23.9 & 82.9 $\pm$ 28.6 & & 3.43 $\pm$ 4.34 & 3.26 $\pm$ 2.65 & & 79.7 $\pm$ 23.3 & 76.1 $\pm$ 27.6 \\
esophagus & 98.4 $\pm$ 3.68 & 97.9 $\pm$ 3.59 & & 1.66 $\pm$ 1.73 & 1.74 $\pm$ 1.10 & & 88.2 $\pm$ 4.03 & 86.96 $\pm$ 4.50 \\
gallbladder & 86.0 $\pm$ 29.1 & 82.8 $\pm$ 33.7 & & 2.71 $\pm$ 3.23 & 3.62 $\pm$ 5.06 & & 78.95 $\pm$ 29.1 & 77.12 $\pm$ 30.1 \\
kidney\_left & 93.3 $\pm$ 17.9 & 91.2 $\pm$ 22.9 & & 3.02 $\pm$ 5.90 & 3.18 $\pm$ 6.42 & & 90.0 $\pm$ 17.4 & 87.2 $\pm$ 22.4 \\
kidney\_right & 94.1 $\pm$ 16.0 & 94.1 $\pm$ 17.8 & & 3.01 $\pm$ 4.79 & 3.72 $\pm$ 11.1 & & 91.3 $\pm$ 15.4 & 91.0 $\pm$ 17.7 \\
liver & 95.6 $\pm$ 12.6 & 92.1 $\pm$ 20.1 & & 3.70 $\pm$ 8.89 & 5.52 $\pm$ 17.6 & & 95.5 $\pm$ 12.0 & 95.1 $\pm$ 12.1 \\
lung\_lower\_lobe\_left & 93.6 $\pm$ 13.9 & 79.6 $\pm$ 34.8 & & 2.62 $\pm$ 2.76 & 4.22 $\pm$ 10.2 & & 92.7 $\pm$ 13.5 & 92.0 $\pm$ 13.3 \\
lung\_lower\_lobe\_right & 90.5 $\pm$ 20.6 & 91.0 $\pm$ 20.0 & & 2.81 $\pm$ 3.16 & 3.03 $\pm$ 5.05 & & 89.7 $\pm$ 21.4 & 92.1 $\pm$ 14.2 \\
lung\_middle\_lobe\_right & 91.5 $\pm$ 10.9 & 90.8 $\pm$ 10.2 & & 4.45 $\pm$ 6.70 & 3.96 $\pm$ 4.45 & & 90.9 $\pm$ 10.2 & 90.5 $\pm$ 9.77 \\
lung\_upper\_lobe\_left & 94.9 $\pm$ 6.74 & 91.4 $\pm$ 16.7 & & 3.08 $\pm$ 3.88 & 6.50 $\pm$ 16.0 & & 93.5 $\pm$ 7.21 & 92.7 $\pm$ 7.17 \\
lung\_upper\_lobe\_right & 87.8 $\pm$ 26.7 & 67.6 $\pm$ 42.6 & & 3.56 $\pm$ 7.73 & 5.77 $\pm$ 17.2 & & 86.9 $\pm$ 26.4 & 86.9 $\pm$ 25.0 \\
pancreas & 88.8 $\pm$ 23.2 & 86.6 $\pm$ 26.2 & & 2.75 $\pm$ 2.65 & 3.74 $\pm$ 6.26 & & 79.96 $\pm$ 23.8 & 78.45 $\pm$ 25.3 \\
prostate & 75.98 $\pm$ 30.1 & 73.53 $\pm$ 33.9 & & 3.47 $\pm$ 2.02 & 3.14 $\pm$ 1.65 & & 77.36 $\pm$ 22.0 & 79.48 $\pm$ 19.6 \\
small\_bowel & 90.0 $\pm$ 14.1 & 85.6 $\pm$ 20.8 & & 4.86 $\pm$ 4.85 & 7.71 $\pm$ 9.50 & & 87.3 $\pm$ 13.2 & 85.6 $\pm$ 14.3 \\
spleen & 98.1 $\pm$ 4.46 & 97.6 $\pm$ 5.03 & & 1.66 $\pm$ 1.60 & 1.86 $\pm$ 1.80 & & 96.5 $\pm$ 1.66 & 96.0 $\pm$ 2.08 \\
stomach & 92.2 $\pm$ 18.9 & 88.0 $\pm$ 24.1 & & 4.78 $\pm$ 8.36 & 5.48 $\pm$ 9.13 & & 90.5 $\pm$ 17.8 & 90.7 $\pm$ 13.7 \\
thyroid\_gland & 97.0 $\pm$ 5.72 & 95.2 $\pm$ 11.1 & & 1.98 $\pm$ 1.00 & 2.69 $\pm$ 3.86 & & 82.8 $\pm$ 8.17 & 80.8 $\pm$ 12.1 \\
trachea & 98.8 $\pm$ 3.36 & 98.6 $\pm$ 3.45 & & 1.65 $\pm$ 2.77 & 1.68 $\pm$ 2.85 & & 90.9 $\pm$ 5.62 & 89.9 $\pm$ 7.00 \\
urinary\_bladder & 91.2 $\pm$ 15.3 & 81.5 $\pm$ 27.2 & & 4.05 $\pm$ 5.09 & 11.2 $\pm$ 25.7 & & 90.4 $\pm$ 13.8 & 86.3 $\pm$ 17.4 \\
mean &  84.5  $\pm 14.9$ & 81.2 $\pm 20.2$ & & 4.27 $\pm 4.76$ & 7.56 $\pm$ 8.58 & & 80.2 $\pm$ 15.2 & 79.7 $\pm$ 16.0
\end{tabular}
\end{adjustbox}
\end{table*}

\begin{table*}[]
\caption{Vertebrae Segmentation Comparison on DeepLabV3+ Backbone}
\centering
\label{tab:vertebra_comparison_deeplabv3}
\begin{adjustbox}{width=0.8\textwidth}
\small
 
    \begin{tabular}{rcccccccc}
    {}                        & \multicolumn{2}{c}{\bfseries{NSD} $\uparrow$}                      &  & \multicolumn{2}{c}{\bfseries{$\text{Haus}_{95} \downarrow $}}                    &  &  \multicolumn{2}{c}{\bfseries{Dice} $\uparrow$}                   \\
    \cmidrule{2-3} \cmidrule{5-6} \cmidrule{8-9}
                              & AIC-Net          & Baseline          &  & AIC-Net            & Baseline           &  & AIC-Net           & Baseline          \\
    sacrum                    & 92.6 $\pm$ 22.1 & 93.2 $\pm$ 18.7 &  & 2.03 $\pm$ 1.51 & 2.34 $\pm$ 2.72 &  & 86.9 $\pm$ 20.7 & 86.7 $\pm$ 17.6 \\
    vertebrae\_C1             & 92.5 $\pm$ 21.0 & 92.6 $\pm$ 21.3 &  & 2.12 $\pm$ 1.23 & 1.99 $\pm$ 1.25 &  & 74.5 $\pm$ 18.8 & 74.2 $\pm$ 19.1 \\
    vertebrae\_C2             & 95.6 $\pm$ 9.15 & 97.2 $\pm$ 6.01 &  & 2.30 $\pm$ 3.44 & 1.63 $\pm$ 0.65 &  & 78.0 $\pm$ 16.4 & 78.3 $\pm$ 14.9 \\
    vertebrae\_C3             & 98.4 $\pm$ 2.08 & 98.5 $\pm$ 1.59 &  & 1.51 $\pm$ 0.41 & 1.64 $\pm$ 0.48 &  & 80.9 $\pm$ 7.85 & 81.1 $\pm$ 5.05 \\
    vertebrae\_C4             & 84.8 $\pm$ 33.7 & 84.9 $\pm$ 33.6 &  & 2.01 $\pm$ 1.90 & 2.00 $\pm$ 1.66 &  & 74.0 $\pm$ 21.6 & 74.2 $\pm$ 21.2 \\
    vertebrae\_C5             & 90.9 $\pm$ 22.7 & 90.4 $\pm$ 23.4 &  & 2.11 $\pm$ 1.55 & 2.41 $\pm$ 2.28 &  & 70.5 $\pm$ 20.7 & 70.9 $\pm$ 21.9 \\
    vertebrae\_C6             & 84.0 $\pm$ 32.5 & 79.3 $\pm$ 38.3 &  & 1.99 $\pm$ 0.96 & 1.90 $\pm$ 1.13 &  & 62.5 $\pm$ 30.7 & 61.0 $\pm$ 33.6 \\
    vertebrae\_C7             & 96.2 $\pm$ 16.1 & 95.7 $\pm$ 16.2 &  & 2.00 $\pm$ 2.87 & 1.97 $\pm$ 2.67 &  & 85.2 $\pm$ 2.16 & 85.0 $\pm$ 2.45 \\
    vertebrae\_L1             & 94.7 $\pm$ 20.5 & 95.5 $\pm$ 18.5 &  & 1.46 $\pm$ 2.01 & 1.47 $\pm$ 2.04 &  & 88.7 $\pm$ 17.4 & 87.8 $\pm$ 19.4 \\
    vertebrae\_L2             & 96.8 $\pm$ 14.1 & 96.5 $\pm$ 15.4 &  & 1.61 $\pm$ 2.19 & 1.57 $\pm$ 2.07 &  & 89.3 $\pm$ 13.7 & 88.5 $\pm$ 16.0 \\
    vertebrae\_L3             & 98.4 $\pm$ 5.08 & 97.7 $\pm$ 8.47 &  & 1.71 $\pm$ 2.59 & 2.05 $\pm$ 3.73 &  & 90.9 $\pm$ 6.55 & 89.8 $\pm$ 9.80 \\
    vertebrae\_L4             & 97.0 $\pm$ 13.9 & 98.4 $\pm$ 7.26 &  & 1.74 $\pm$ 2.83 & 1.81 $\pm$ 3.12 &  & 88.7 $\pm$ 13.8 & 88.9 $\pm$ 10.0 \\
    vertebrae\_L5             & 98.4 $\pm$ 5.71 & 98.5 $\pm$ 6.23 &  & 1.71 $\pm$ 2.57 & 1.57 $\pm$ 2.24 &  & 89.9 $\pm$ 7.33 & 89.6 $\pm$ 8.42 \\
    vertebrae\_S1             & 95.1 $\pm$ 19.3 & 96.4 $\pm$ 14.5 &  & 1.41 $\pm$ 1.67 & 1.46 $\pm$ 1.67 &  & 88.4 $\pm$ 18.2 & 88.8 $\pm$ 14.5 \\
    vertebrae\_T1             & 99.3 $\pm$ 1.93 & 99.3 $\pm$ 1.86 &  & 1.49 $\pm$ 1.88 & 1.31 $\pm$ 1.02 &  & 89.0 $\pm$ 1.88 & 88.5 $\pm$ 2.58 \\
    vertebrae\_T10            & 95.0 $\pm$ 16.6 & 95.9 $\pm$ 15.6 &  & 2.11 $\pm$ 3.08 & 1.85 $\pm$ 2.64 &  & 85.6 $\pm$ 20.3 & 86.1 $\pm$ 18.5 \\
    vertebrae\_T11            & 94.3 $\pm$ 18.6 & 96.9 $\pm$ 13.3 &  & 1.82 $\pm$ 2.53 & 1.63 $\pm$ 2.39 &  & 87.6 $\pm$ 15.4 & 88.3 $\pm$ 14.9 \\
    vertebrae\_T12            & 95.0 $\pm$ 19.8 & 96.3 $\pm$ 16.2 &  & 2.25 $\pm$ 5.16 & 1.78 $\pm$ 3.31 &  & 87.6 $\pm$ 19.6 & 88.2 $\pm$ 16.6 \\
    vertebrae\_T2             & 98.5 $\pm$ 5.21 & 97.8 $\pm$ 8.67 &  & 1.65 $\pm$ 1.95 & 1.44 $\pm$ 1.67 &  & 88.0 $\pm$ 6.79 & 87.0 $\pm$ 11.3 \\
    vertebrae\_T3             & 97.2 $\pm$ 10.7 & 96.6 $\pm$ 12.8 &  & 1.96 $\pm$ 2.24 & 1.71 $\pm$ 2.33 &  & 86.8 $\pm$ 11.9 & 85.4 $\pm$ 15.4 \\
    vertebrae\_T4             & 96.2 $\pm$ 15.8 & 96.1 $\pm$ 16.1 &  & 1.63 $\pm$ 1.80 & 1.63 $\pm$ 2.01 &  & 85.7 $\pm$ 15.8 & 85.4 $\pm$ 16.0 \\
    vertebrae\_T5             & 93.7 $\pm$ 21.4 & 95.2 $\pm$ 16.0 &  & 1.72 $\pm$ 1.93 & 2.20 $\pm$ 2.55 &  & 84.7 $\pm$ 15.6 & 83.7 $\pm$ 16.0 \\
    vertebrae\_T6             & 89.9 $\pm$ 25.7 & 91.2 $\pm$ 22.9 &  & 1.93 $\pm$ 2.03 & 2.50 $\pm$ 4.74 &  & 81.3 $\pm$ 20.6 & 79.9 $\pm$ 22.6 \\
    vertebrae\_T7             & 86.4 $\pm$ 29.2 & 85.0 $\pm$ 30.2 &  & 3.59 $\pm$ 6.51 & 3.85 $\pm$ 6.88 &  & 75.0 $\pm$ 28.4 & 74.1 $\pm$ 29.6 \\
    vertebrae\_T8             & 89.6 $\pm$ 25.1 & 90.1 $\pm$ 25.3 &  & 3.11 $\pm$ 3.88 & 2.73 $\pm$ 3.62 &  & 78.3 $\pm$ 25.0 & 79.0 $\pm$ 25.6 \\
    vertebrae\_T9             & 94.8 $\pm$ 17.0 & 95.2 $\pm$ 17.2 &  & 2.27 $\pm$ 3.13 & 1.95 $\pm$ 2.71 &  & 84.8 $\pm$ 20.1 & 84.6 $\pm$ 20.3 \\
    mean                      & 94.1 $\pm$ 17.1 & 94.1 $\pm$ 16.4 &  & 1.94 $\pm$ 2.46 & 1.94 $\pm$ 2.56 &  & 83.2 $\pm$ 16.0 & 82.9 $\pm$ 16.3
    \end{tabular}
\end{adjustbox}
\end{table*}

\begin{table*}[]
\caption{Organ Segmentation Comparison on UNETR Backbone}
\centering
\label{tab:organ_comparison_unetr}
\begin{adjustbox}{width=0.8\textwidth}
\small
 
\begin{tabular}{rcccccccc}
{} & \multicolumn{2}{c}{\bfseries{NSD} $\uparrow$} & & \multicolumn{2}{c}{\bfseries{$\text{Haus}_{95} \downarrow $}} & & \multicolumn{2}{c}{\bfseries{Dice} $\uparrow$} \\
\cmidrule{2-3} \cmidrule{5-6} \cmidrule{8-9}
& AIC-Net & Baseline & & AIC-Net & Baseline & & AIC-Net & Baseline \\
adrenal\_gland\_left & 87.0 $\pm$ 23.3 & 78.0 $\pm$ 28.7 & & 4.36 $\pm$ 4.36 & 10.8 $\pm$ 28.0 & & 76.5 $\pm$ 19.0 & 69.5 $\pm$ 20.5 \\
adrenal\_gland\_right & 92.8 $\pm$ 18.1 & 88.9 $\pm$ 21.8 & & 2.60 $\pm$ 3.70 & 3.46 $\pm$ 4.60 & & 78.0 $\pm$ 16.1 & 73.1 $\pm$ 19.7 \\
colon & 61.7 $\pm$ 29.6 & 49.7 $\pm$ 26.6 & & 20.1 $\pm$ 16.9 & 37.6 $\pm$ 28.6 & & 77.4 $\pm$ 12.5 & 69.5 $\pm$ 13.9 \\
duodenum & 69.5 $\pm$ 23.9 & 51.2 $\pm$ 26.7 & & 9.53 $\pm$ 12.0 & 25.9 $\pm$ 39.8 & & 66.9 $\pm$ 23.3 & 57.1 $\pm$ 23.2 \\
esophagus & 83.9 $\pm$ 28.7 & 73.9 $\pm$ 32.5 & & 5.00 $\pm$ 12.3 & 22.0 $\pm$ 49.8 & & 81.9 $\pm$ 10.5 & 75.5 $\pm$ 13.5 \\
gallbladder & 62.1 $\pm$ 40.6 & 55.0 $\pm$ 39.1 & & 9.31 $\pm$ 13.1 & 13.5 $\pm$ 13.2 & & 70.4 $\pm$ 30.6 & 66.0 $\pm$ 29.9 \\
kidney\_left & 83.6 $\pm$ 28.6 & 75.5 $\pm$ 32.7 & & 6.84 $\pm$ 15.9 & 14.4 $\pm$ 29.5 & & 86.3 $\pm$ 22.6 & 82.7 $\pm$ 23.2 \\
kidney\_right & 81.8 $\pm$ 32.7 & 73.8 $\pm$ 35.1 & & 6.16 $\pm$ 10.1 & 23.6 $\pm$ 38.7 & & 89.5 $\pm$ 18.6 & 85.8 $\pm$ 20.0 \\
liver & 86.1 $\pm$ 22.2 & 76.6 $\pm$ 28.1 & & 15.3 $\pm$ 61.7 & 17.4 $\pm$ 37.1 & & 93.9 $\pm$ 12.4 & 91.6 $\pm$ 14.2 \\
lung\_lower\_lobe\_left & 81.5 $\pm$ 26.0 & 76.9 $\pm$ 28.9 & & 18.3 $\pm$ 56.2 & 22.8 $\pm$ 44.9 & & 88.5 $\pm$ 17.7 & 88.2 $\pm$ 16.1 \\
lung\_lower\_lobe\_right & 84.6 $\pm$ 22.5 & 78.3 $\pm$ 28.3 & & 13.9 $\pm$ 33.4 & 19.8 $\pm$ 44.7 & & 90.1 $\pm$ 15.1 & 89.6 $\pm$ 13.8 \\
lung\_middle\_lobe\_right & 76.8 $\pm$ 23.4 & 75.1 $\pm$ 20.6 & & 7.24 $\pm$ 7.35 & 14.4 $\pm$ 27.9 & & 85.9 $\pm$ 12.3 & 84.4 $\pm$ 12.1 \\
lung\_upper\_lobe\_left & 78.0 $\pm$ 28.8 & 69.5 $\pm$ 32.5 & & 11.2 $\pm$ 26.0 & 32.0 $\pm$ 53.6 & & 88.9 $\pm$ 12.9 & 86.6 $\pm$ 14.9 \\
lung\_upper\_lobe\_right & 62.1 $\pm$ 41.2 & 52.3 $\pm$ 43.1 & & 10.9 $\pm$ 26.5 & 27.5 $\pm$ 52.3 & & 84.3 $\pm$ 25.8 & 83.2 $\pm$ 26.8 \\
pancreas & 74.4 $\pm$ 26.6 & 63.7 $\pm$ 24.3 & & 7.70 $\pm$ 7.37 & 12.7 $\pm$ 15.7 & & 72.1 $\pm$ 24.3 & 64.0 $\pm$ 24.3 \\
prostate & 36.5 $\pm$ 39.5 & 32.2 $\pm$ 35.7 & & 5.02 $\pm$ 3.55 & 9.45 $\pm$ 16.7 & & 73.2 $\pm$ 16.0 & 68.9 $\pm$ 18.6 \\
small\_bowel & 65.2 $\pm$ 28.6 & 51.8 $\pm$ 28.7 & & 17.6 $\pm$ 14.4 & 29.7 $\pm$ 23.5 & & 75.4 $\pm$ 16.1 & 67.5 $\pm$ 15.5 \\
spleen & 89.8 $\pm$ 18.4 & 77.4 $\pm$ 29.4 & & 10.2 $\pm$ 26.4 & 14.8 $\pm$ 23.1 & & 94.8 $\pm$ 4.14 & 92.2 $\pm$ 5.46 \\
stomach & 77.4 $\pm$ 26.8 & 61.7 $\pm$ 29.1 & & 14.5 $\pm$ 17.9 & 27.4 $\pm$ 28.4 & & 84.8 $\pm$ 20.5 & 78.9 $\pm$ 19.5 \\
thyroid\_gland & 76.4 $\pm$ 35.7 & 47.2 $\pm$ 39.6 & & 6.96 $\pm$ 25.8 & 46.7 $\pm$ 70.0 & & 80.1 $\pm$ 9.45 & 69.6 $\pm$ 10.5 \\
trachea & 79.4 $\pm$ 36.8 & 67.3 $\pm$ 41.9 & & 7.21 $\pm$ 30.9 & 24.5 $\pm$ 57.9 & & 89.7 $\pm$ 7.89 & 87.6 $\pm$ 8.79 \\
urinary\_bladder & 66.1 $\pm$ 28.1 & 48.4 $\pm$ 30.3 & & 14.1 $\pm$ 25.3 & 34.2 $\pm$ 50.1 & & 79.1 $\pm$ 20.2 & 73.9 $\pm$ 20.2 \\
mean & 74.4 $\pm$ 34.7  & 53.2 $\pm$ 39.8  & & 12.6 $\pm$ 31.2 & 52.2 $\pm$ 69.8 & & 83.2 $\pm$ 20.2 & 76.5 $\pm$ 20.7 
\end{tabular}
\end{adjustbox}
\end{table*}

\begin{table*}[]
\caption{Vertebrae Segmentation Comparison on UNETR Backbone}
\centering
\label{tab:vertebra_comparison_unetrvit}
\begin{adjustbox}{width=0.8\textwidth}
\small
 
    \begin{tabular}{rcccccccc}
    {}                        & \multicolumn{2}{c}{\bfseries{NSD} $\uparrow$}                      &  & \multicolumn{2}{c}{\bfseries{$\text{Haus}_{95} \downarrow $}}                    &  &  \multicolumn{2}{c}{\bfseries{Dice} $\uparrow$}                   \\
    \cmidrule{2-3} \cmidrule{5-6} \cmidrule{8-9}
                              & AIC-Net          & Baseline          &  & AIC-Net            & Baseline           &  & AIC-Net           & Baseline          \\
    sacrum                    & 67.6 $\pm$ 42.7  & 55.7 $\pm$ 39.3  &  & 20.5 $\pm$ 47.1   & 68.2 $\pm$ 43.1   &  & 86.0 $\pm$ 21.1   & 79.9 $\pm$ 19.6  \\
    vertebrae\_C1             & 27.8 $\pm$ 42.9  & 20.6 $\pm$ 36.7  &  & 53.0 $\pm$ 109    & 85.7 $\pm$ 126    &  & 78.3 $\pm$ 20.1   & 66.7 $\pm$ 23.3  \\
    vertebrae\_C2             & 39.0 $\pm$ 47.3  & 23.4 $\pm$ 38.9  &  & 25.2 $\pm$ 81.2   & 54.7 $\pm$ 106    &  & 83.8 $\pm$ 12.4   & 75.2 $\pm$ 16.9  \\
    vertebrae\_C3             & 42.0 $\pm$ 48.4  & 22.4 $\pm$ 39.7  &  & 51.8 $\pm$ 127    & 48.2 $\pm$ 99.1   &  & 86.0 $\pm$ 10.0   & 80.7 $\pm$ 16.2  \\
    vertebrae\_C4             & 47.0 $\pm$ 48.4  & 20.6 $\pm$ 38.8  &  & 48.7 $\pm$ 119    & 74.9 $\pm$ 127    &  & 78.8 $\pm$ 25.6   & 76.4 $\pm$ 24.0  \\
    vertebrae\_C5             & 70.8 $\pm$ 43.1  & 35.5 $\pm$ 46.1  &  & 18.4 $\pm$ 74.6   & 43.1 $\pm$ 104    &  & 84.5 $\pm$ 11.5   & 77.2 $\pm$ 19.9  \\
    vertebrae\_C6             & 72.3 $\pm$ 40.3  & 44.4 $\pm$ 44.1  &  & 4.23 $\pm$ 13.1   & 77.7 $\pm$ 117    &  & 75.3 $\pm$ 24.4   & 64.1 $\pm$ 30.6  \\
    vertebrae\_C7             & 72.4 $\pm$ 42.6  & 58.6 $\pm$ 44.6  &  & 8.65 $\pm$ 27.1   & 67.0 $\pm$ 117    &  & 89.5 $\pm$ 8.93   & 83.4 $\pm$ 11.4  \\
    vertebrae\_L1             & 87.3 $\pm$ 28.2  & 66.7 $\pm$ 38.5  &  & 4.14 $\pm$ 7.78   & 37.6 $\pm$ 59.0   &  & 87.1 $\pm$ 22.8   & 80.5 $\pm$ 22.5  \\
    vertebrae\_L2             & 85.1 $\pm$ 30.8  & 63.6 $\pm$ 38.9  &  & 4.04 $\pm$ 9.38   & 42.4 $\pm$ 55.9   &  & 88.8 $\pm$ 18.5   & 80.8 $\pm$ 19.0  \\
    vertebrae\_L3             & 88.1 $\pm$ 26.7  & 61.9 $\pm$ 40.1  &  & 7.44 $\pm$ 30.9   & 77.8 $\pm$ 70.3   &  & 90.5 $\pm$ 15.8   & 83.4 $\pm$ 17.3  \\
    vertebrae\_L4             & 85.5 $\pm$ 31.5  & 66.2 $\pm$ 40.1  &  & 10.4 $\pm$ 34.9   & 60.8 $\pm$ 68.7   &  & 89.0 $\pm$ 20.2   & 83.9 $\pm$ 20.9  \\
    vertebrae\_L5             & 93.6 $\pm$ 19.0  & 67.0 $\pm$ 39.2  &  & 7.13 $\pm$ 19.9   & 59.5 $\pm$ 61.4   &  & 92.1 $\pm$ 9.37   & 86.6 $\pm$ 12.6  \\
    vertebrae\_S1             & 89.8 $\pm$ 27.0  & 65.8 $\pm$ 41.0  &  & 7.54 $\pm$ 25.3   & 47.3 $\pm$ 48.4   &  & 88.1 $\pm$ 17.2   & 84.0 $\pm$ 17.2  \\
    vertebrae\_T1             & 84.2 $\pm$ 34.2  & 51.7 $\pm$ 46.7  &  & 2.20 $\pm$ 2.97   & 36.4 $\pm$ 80.6   &  & 91.3 $\pm$ 7.09   & 86.3 $\pm$ 9.66  \\
    vertebrae\_T10            & 84.6 $\pm$ 28.2  & 62.5 $\pm$ 36.3  &  & 4.15 $\pm$ 4.91   & 55.6 $\pm$ 58.6   &  & 84.5 $\pm$ 22.5   & 75.3 $\pm$ 22.3  \\
    vertebrae\_T11            & 84.5 $\pm$ 31.6  & 64.1 $\pm$ 38.9  &  & 2.88 $\pm$ 4.09   & 38.7 $\pm$ 56.1   &  & 86.1 $\pm$ 24.2   & 81.2 $\pm$ 18.3  \\
    vertebrae\_T12            & 86.9 $\pm$ 29.5  & 71.0 $\pm$ 37.6  &  & 3.48 $\pm$ 6.80   & 25.9 $\pm$ 58.2   &  & 86.0 $\pm$ 25.1   & 81.7 $\pm$ 23.7  \\
    vertebrae\_T2             & 89.7 $\pm$ 24.1  & 58.2 $\pm$ 44.3  &  & 4.65 $\pm$ 9.75   & 22.7 $\pm$ 28.3   &  & 88.6 $\pm$ 14.8   & 83.5 $\pm$ 15.4  \\
    vertebrae\_T3             & 88.8 $\pm$ 23.7  & 58.1 $\pm$ 42.6  &  & 3.23 $\pm$ 3.45   & 41.3 $\pm$ 45.5   &  & 86.2 $\pm$ 16.4   & 80.3 $\pm$ 17.5  \\
    vertebrae\_T4             & 79.7 $\pm$ 34.3  & 57.9 $\pm$ 40.9  &  & 3.55 $\pm$ 3.45   & 51.2 $\pm$ 47.3   &  & 80.7 $\pm$ 25.5   & 73.4 $\pm$ 24.1  \\
    vertebrae\_T5             & 79.0 $\pm$ 30.9  & 56.0 $\pm$ 38.9  &  & 5.28 $\pm$ 6.26   & 41.8 $\pm$ 49.7   &  & 76.5 $\pm$ 25.2   & 66.7 $\pm$ 28.1  \\
    vertebrae\_T6             & 71.6 $\pm$ 36.3  & 56.4 $\pm$ 36.7  &  & 5.83 $\pm$ 8.92   & 46.1 $\pm$ 39.8   &  & 68.0 $\pm$ 35.4   & 62.8 $\pm$ 28.5  \\
    vertebrae\_T7             & 69.1 $\pm$ 39.0  & 55.6 $\pm$ 35.5  &  & 5.38 $\pm$ 6.67   & 62.2 $\pm$ 47.5   &  & 66.2 $\pm$ 35.5   & 62.6 $\pm$ 26.7  \\
    vertebrae\_T8             & 71.6 $\pm$ 37.8  & 56.9 $\pm$ 35.8  &  & 10.7 $\pm$ 31.5   & 45.2 $\pm$ 48.7   &  & 73.0 $\pm$ 30.3   & 61.5 $\pm$ 30.1  \\
    vertebrae\_T9             & 77.6 $\pm$ 34.1  & 62.8 $\pm$ 34.8  &  & 4.54 $\pm$ 5.34   & 45.9 $\pm$ 50.4   &  & 79.6 $\pm$ 26.4   & 71.0 $\pm$ 23.0  \\
    mean                      & 74.4 $\pm$ 34.7  & 53.2 $\pm$ 39.8  &  & 12.6 $\pm$ 31.5   & 52.2 $\pm$ 69.7   &  & 83.2 $\pm$ 20.2   & 76.5 $\pm$ 20.7  \\
    \end{tabular}
\end{adjustbox}
\end{table*}

\begin{table*}[]
\caption{Organ Segmentation Comparison on UNETR-Swin Backbone}
\centering
\label{tab:organ_comparison_unetr_swin}
\begin{adjustbox}{width=0.8\textwidth}
\small
 
    \begin{tabular}{rcccccccc}
    {}                        & \multicolumn{2}{c}{\bfseries{NSD} $\uparrow$}                      &  & \multicolumn{2}{c}{\bfseries{$\text{Haus}_{95} \downarrow $}}                    &  &  \multicolumn{2}{c}{\bfseries{Dice} $\uparrow$}                   \\
    \cmidrule{2-3} \cmidrule{5-6} \cmidrule{8-9}
                              & AIC-Net          & Baseline          &  & AIC-Net            & Baseline           &  & AIC-Net           & Baseline          \\
    adrenal\_gland\_left      & 95.0 $\pm$  13.9  & 96.2 $\pm$  10.2 &  & 2.22  $\pm$  2.34  & 5.29 $\pm$  26.0 &  & 82.2 $\pm$  15.2  & 83.4 $\pm$  15.3 \\
    adrenal\_gland\_right     & 96.9 $\pm$  9.41 & 96.8 $\pm$  9.50 &  & 1.65  $\pm$  1.47  & 1.55 $\pm$  1.46  &  & 83.2 $\pm$  14.1  & 83.7 $\pm$  15.2 \\
    colon                     & 87.1 $\pm$  15.5 & 82.0  $\pm$  25.1 &  & 10.6 $\pm$  13.7 & 12.6$\pm$ 17.6  &  & 86.1  $\pm$  11.5 & 85.1 $\pm$  15.6 \\
    duodenum                  & 86.9 $\pm$  16.1 & 85.0  $\pm$  22.6 &  & 5.56  $\pm$  6.47  & 8.02 $\pm$ 17.4  &  & 78.2  $\pm$  18.9 & 78.6 $\pm$  21.0  \\
    esophagus                 & 96.1 $\pm$  13.1 & 95.7 $\pm$  13.3 &  & 2.82  $\pm$  6.06  & 2.69 $\pm$  6.04   &  & 89.5 $\pm$  5.70  & 89.0  $\pm$  6.59 \\
    gallbladder               & 74.2 $\pm$  38.0  & 78.1 $\pm$  35.0  &  & 7.12  $\pm$  16.8  & 6.12 $\pm$  17.5 &  & 80.2 $\pm$  25.3  & 81.1 $\pm$  24.0  \\
    kidney\_left              & 94.0  $\pm$  14.8 & 91.5 $\pm$  21.9 &  & 5.05  $\pm$ 13.2 & 4.68   $\pm$  13.2 &  & 91.2 $\pm$  15.6  & 91.3 $\pm$  16.8 \\
    kidney\_right             & 90.1 $\pm$  25.2 & 89.7 $\pm$  25.8 &  & 3.63  $\pm$  6.58  & 3.67 $\pm$  7.35  &  & 92.2 $\pm$  16.1  & 91.8 $\pm$  17.7 \\
    liver                     & 92.9 $\pm$  17.3 & 87.9 $\pm$  27.0  &  & 6.40  $\pm$  14.7 & 9.30 $\pm$  30.8 &  & 95.0  $\pm$  12.3  & 95.2 $\pm$  12.3 \\
    lung\_lower\_lobe\_left   & 93.2 $\pm$  11.1 & 87.4 $\pm$  25.1 &  & 2.92  $\pm$  3.12  & 7.85 $\pm$  25.9 &  & 92.9 $\pm$  12.7  & 92.7 $\pm$  13.3 \\
    lung\_lower\_lobe\_right  & 90.4 $\pm$  20.0   & 86.5 $\pm$  26.3 &  & 2.90  $\pm$  3.38   & 10.8 $\pm$  37.5 &  & 92.3 $\pm$  14.3  & 92.4 $\pm$  14.6 \\
    lung\_middle\_lobe\_right & 88.5 $\pm$  15.5 & 84.5 $\pm$  23.9 &  & 4.37  $\pm$ 4.96  & 6.85  $\pm$  18.0 &  & 90.6 $\pm$  9.94  & 90.6 $\pm$  9.70 \\
    lung\_upper\_lobe\_left   & 91.1 $\pm$  17.1 & 90.0   $\pm$  20.0   &  & 3.95  $\pm$  7.28  & 5.40 $\pm$  14.1 &  & 93.5 $\pm$  6.86  & 93.6 $\pm$  6.03  \\
    lung\_upper\_lobe\_right  & 72.7 $\pm$  38.4 & 64.5 $\pm$  43.6 &  & 5.17   $\pm$  9.24  & 5.48 $\pm$  9.83  &  & 87.3 $\pm$  22.4  & 87.3 $\pm$  25.1 \\
    pancreas                  & 88.1 $\pm$  22.3 & 88.6 $\pm$  21.5 &  & 3.57  $\pm$  3.58  & 5.29 $\pm$  13.5 &  & 82.7 $\pm$  19.4  & 83.3 $\pm$  19.1 \\
    prostate                  & 44.4 $\pm$  45.5 & 40.3 $\pm$  43.1 &  & 3.75  $\pm$  3.36  & 4.60 $\pm$  6.07  &  & 79.1 $\pm$  20.6  & 79.6 $\pm$  13.4 \\
    small\_bowel              & 82.6 $\pm$  22.9 & 83.5 $\pm$  23.0  &  & 12.6 $\pm$  17.7 & 12.5 $\pm$  25.9 &  & 84.8 $\pm$  12.9  & 85.7 $\pm$  13.0  \\
    spleen                    & 95.8 $\pm$  12.9 & 91.1 $\pm$  23.4 &  & 3.15  $\pm$  6.72  & 6.81 $\pm$  22.4 &  & 96.6 $\pm$  2.36  & 96.6 $\pm$  18.6 \\
    stomach                   & 89.6 $\pm$  20.1 & 82.0  $\pm$  30.8 &  & 7.59  $\pm$  14.1 & 11.8 $\pm$  19.3 &  & 90.2 $\pm$  15.8  & 90.5 $\pm$  15.0  \\
    thyroid\_gland            & 92.7 $\pm$  21.6 & 70.2 $\pm$  44.3 &  & 5.74  $\pm$  16.9 & 4.80 $\pm$  19.6 &  & 87.3 $\pm$  8.14  & 87.1 $\pm$  11.0  \\
    trachea                   & 89.9 $\pm$  27.9 & 82.8 $\pm$  36.2 &  & 7.56   $\pm$  34.8 & 5.24 $\pm$  17.8 &  & 92.8 $\pm$  5.40  & 92.2 $\pm$  6.35 \\
    urinary\_bladder          & 83.2 $\pm$  24.3 & 75.0  $\pm$  32.5 &  & 9.52  $\pm$  21.9 & 17.0  $\pm$  32.5 &  & 87.2 $\pm$  15.5  & 86.4 $\pm$  15.9 \\
    mean                      & 80.4 $\pm$  21.1 & 76.7  $\pm$  25.7 &  & 6.18  $\pm$ 11.9  & 7.89  $\pm$ 17.9  &  & 84.2  $\pm$ 15.9      & 84.1 $\pm$ 15.4
    \end{tabular}
\end{adjustbox}
\end{table*}

\begin{table*}[]
\caption{Vertebrae Segmentation Comparison on UNETR-Swin Backbone}
\centering
\label{tab:vertebra_comparison_unetr_swin}
\begin{adjustbox}{width=0.8\textwidth}
\small
 
    \begin{tabular}{rcccccccc}
    {}                        & \multicolumn{2}{c}{\bfseries{NSD $\uparrow$}}                      &  & \multicolumn{2}{c}{\bfseries{$\text{Haus}_{95} \downarrow$}}                    &  &  \multicolumn{2}{c}{\bfseries{Dice} $\uparrow$}                   \\
    \cmidrule{2-3} \cmidrule{5-6} \cmidrule{8-9}
                              & AIC-Net          & Baseline          &  & AIC-Net            & Baseline           &  & AIC-Net           & Baseline          \\
sacrum         & 92.9 $\pm$ 22.1 & 78.2 $\pm$ 38.7 &  & 1.45  $\pm$ 0.56 & 16.6 $\pm$ 44.0 &  & 89.3 $\pm$ 21.3 & 90.0   $\pm$ 19.2 \\
vertebrae\_C1  & 93.6 $\pm$ 21.0  & 41.0  $\pm$ 48.7 &  & 1.48  $\pm$ 0.77 & 12.7 $\pm$ 51.5 &  & 83.6 $\pm$ 20.2 & 84.1 $\pm$ 20.9 \\
vertebrae\_C2  & 97.3 $\pm$ 6.00  & 58.9 $\pm$ 48.9 &  & 2.04 $\pm$ 3.23 & 1.37 $\pm$ 0.71  &  & 86.1 $\pm$ 14.2 & 87.6 $\pm$ 12.8 \\
vertebrae\_C3  & 97.7 $\pm$ 7.06 & 66.0  $\pm$ 47.9 &  & 1.29 $\pm$ 0.57 & 1.24 $\pm$ 0.49  &  & 86.8 $\pm$ 17.0  & 90.3 $\pm$ 8.12 \\
vertebrae\_C4  & 91.9 $\pm$ 25.6 & 56.7 $\pm$ 49.1 &  & 1.36 $\pm$ 0.64 & 9.03 $\pm$ 27.8  &  & 83.9 $\pm$ 23.8 & 83.4 $\pm$ 24.1 \\
vertebrae\_C5  & 93.4 $\pm$ 22.2 & 70.5 $\pm$ 43.1 &  & 1.38 $\pm$ 0.96 & 1.81 $\pm$ 1.49  &  & 83.8 $\pm$ 21.0  & 84.9 $\pm$ 14.7 \\
vertebrae\_C6  & 92.5 $\pm$ 24.3 & 82.7 $\pm$ 35.2 &  & 1.15 $\pm$ 0.46 & 17.9 $\pm$ 64.8 &  & 79.9 $\pm$ 25.3 & 82.4 $\pm$ 20.9 \\
vertebrae\_C7  & 96.2 $\pm$ 16.4 & 74.8 $\pm$ 43.1 &  & 1.42 $\pm$ 1.68 & 3.53 $\pm$ 14.5 &  & 92.8 $\pm$ 2.90 & 93.8 $\pm$ 1.81 \\
vertebrae\_L1  & 93.6 $\pm$ 22.1 & 92.7 $\pm$ 24.3 &  & 1.74 $\pm$ 2.83 & 2.48 $\pm$ 6.98  &  & 90.9 $\pm$ 20.4 & 93.3 $\pm$ 15.5 \\
vertebrae\_L2  & 96.6 $\pm$ 14.0  & 90.8 $\pm$ 27.0  &  & 1.73 $\pm$ 2.64 & 2.77 $\pm$ 8.28  &  & 92.8 $\pm$ 13.6 & 95.1 $\pm$ 5.79 \\
vertebrae\_L3  & 98.4 $\pm$ 4.97  & 92.4 $\pm$ 24.8 &  & 1.92 $\pm$ 2.82 & 1.44 $\pm$ 1.77   &  & 94.6 $\pm$ 6.20 & 95.7 $\pm$ 3.00  \\
vertebrae\_L4  & 97.1 $\pm$ 13.5 & 92.5 $\pm$ 24.4 &  & 1.57 $\pm$ 2.40 & 8.77 $\pm$ 42.5 &  & 93.2 $\pm$ 13.3 & 94.9 $\pm$ 6.19 \\
vertebrae\_L5  & 99.0  $\pm$ 2.98  & 97.4 $\pm$ 13.1 &  & 1.49 $\pm$ 2.10 & 1.22  $\pm$ 0.72  &  & 94.7 $\pm$ 3.78 & 95.4 $\pm$ 2.42 \\
vertebrae\_S1  & 93.5 $\pm$ 22.8 & 92.0  $\pm$ 25.4 &  & 1.35 $\pm$ 1.36 & 1.32 $\pm$ 0.93  &  & 89.8 $\pm$ 18.0  & 91.3 $\pm$ 13.1 \\
vertebrae\_T1  & 99.5 $\pm$ 1.45 & 64.9 $\pm$ 47.5 &  & 1.30 $\pm$ 1.22 & 12.4 $\pm$ 33.6 &  & 94.0  $\pm$ 23.0 & 94.6 $\pm$ 1.69 \\
vertebrae\_T10 & 96.1 $\pm$ 15.0  & 89.3 $\pm$ 28.2 &  & 1.78 $\pm$ 2.66 & 2.00 $\pm$ 3.06   &  & 91.4 $\pm$ 17.7 & 90.6 $\pm$ 20.1 \\
vertebrae\_T11 & 95.3 $\pm$ 17.9 & 88.4 $\pm$ 30.3 &  & 1.70   $\pm$ 2.55 & 1.61 $\pm$ 2.26  &  & 92.6 $\pm$ 14.7 & 93.1 $\pm$ 14.3 \\
vertebrae\_T12 & 96.4 $\pm$ 15.6 & 90.0   $\pm$ 28.5 &  & 1.92 $\pm$ 3.26 & 2.08 $\pm$ 3.78  &  & 92.3 $\pm$ 17.1 & 92.1 $\pm$ 19.0  \\
vertebrae\_T2  & 98.5 $\pm$ 5.17 & 74.7 $\pm$ 42.2 &  & 1.68 $\pm$ 1.90 & 8.49 $\pm$ 27.3 &  & 93.0  $\pm$ 6.70 & 93.5 $\pm$ 7.21 \\
vertebrae\_T3  & 97.0  $\pm$ 11.0  & 74.0  $\pm$ 42.8 &  & 2.05 $\pm$ 2.56  & 14.2 $\pm$ 39.0 &  & 91.8 $\pm$ 12.5 & 92.3 $\pm$ 13.6 \\
vertebrae\_T4  & 96.0  $\pm$ 16.1 & 73.7 $\pm$ 42.2 &  & 1.67 $\pm$ 2.10 & 19.4 $\pm$ 49.4  &  & 90.0   $\pm$ 16.6 & 90.1 $\pm$ 18.0  \\
vertebrae\_T5  & 95.0  $\pm$ 16.4 & 72.7 $\pm$ 42.7 &  & 2.27 $\pm$ 2.99 & 16.2 $\pm$ 41.7 &  & 89.5 $\pm$ 15.9 & 90.3 $\pm$ 15.8 \\
vertebrae\_T6  & 94.5 $\pm$ 16.3 & 80.3 $\pm$ 37.7 &  & 2.03 $\pm$ 2.24 & 1.62 $\pm$ 1.68  &  & 86.1 $\pm$ 20.9 & 87.7 $\pm$ 21.1 \\
vertebrae\_T7  & 87.7 $\pm$ 28.2 & 86.4 $\pm$ 30.4 &  & 3.57 $\pm$ 7.08 & 3.72 $\pm$ 7.09  &  & 81.9 $\pm$ 28.1 & 83.8 $\pm$ 26.6 \\
vertebrae\_T8  & 92.4 $\pm$ 22.8 & 88.9 $\pm$ 28.3 &  & 2.71 $\pm$ 4.67 & 5.33 $\pm$ 22.6 &  & 86.8 $\pm$ 23.9 & 85.9 $\pm$ 24.8 \\
vertebrae\_T9  & 95.3 $\pm$ 17.6 & 90.3 $\pm$ 26.1 &  & 1.71 $\pm$ 2.36 & 2.11 $\pm$ 3.01  &  & 90.6 $\pm$ 20.1 & 89.9 $\pm$ 20.2 \\
mean           & 95.3 $\pm$ 15.7 & 79.2 $\pm$ 35.4 &  & 1.76 $\pm$ 2.25 & 6.59 $\pm$ 19.3  &  & 89.3 $\pm$ 16.1 & 90.2 $\pm$ 14.3
    \end{tabular}
\end{adjustbox}
\end{table*}

\end{document}